\def\year{2020}\relax
\pdfoutput=1
\documentclass[leqno,onefignum,onetabnum]{siamltex}
\usepackage[left=2.4cm,right=2.4cm, top=2cm, bottom=2cm]{geometry}

\usepackage{times}
\usepackage{helvet}
\usepackage{courier}
\usepackage{xspace}
\usepackage[toc,page]{appendix}
\usepackage[ruled,noend]{algorithm2e}

\SetCommentSty{mycommfont}
\SetKwInput{KwInput}{Input}

\usepackage[font=scriptsize]{subfig}
 \usepackage[dvipsnames]{xcolor}

\usepackage{amsmath,amsfonts,bm}

\usepackage{xcolor}
\usepackage{colortbl}

\def\eqref#1{equation~\ref{#1}}

\def\1{\bm{1}}

\DeclareMathAlphabet{\mathsfit}{\encodingdefault}{\sfdefault}{m}{sl}
\SetMathAlphabet{\mathsfit}{bold}{\encodingdefault}{\sfdefault}{bx}{n}

\usepackage{booktabs} 
\usepackage{multicol}
\usepackage{diagbox}

\usepackage[utf8]{inputenc} %
\usepackage[T1]{fontenc}    %
\usepackage{url}            %
\usepackage{booktabs}       %
\usepackage{amsfonts}       %
\usepackage{nicefrac}       %
\usepackage{microtype}      %
\usepackage{color}
\usepackage{amsmath}
\usepackage{amsthm}
\usepackage{graphicx}
\usepackage{tikz}
\usepackage{caption}
\usepackage{multirow}
\usepackage{wrapfig}
\usepackage{amsfonts, amsmath}
\usepackage{enumitem}
\usepackage{bbding}
\usepackage{pifont}
\usepackage{wasysym}
\usepackage{amssymb}
\usepackage{soul}
\usepackage{bm}
\usepackage{comment}
\usepackage{wrapfig}
\usepackage{listings}

\usepackage{siunitx}

\definecolor{light-gray}{gray}{0.80}

\renewcommand\paragraph{\subsubsection*}

\newcommand\eref{Eq.~\ref}
\newcommand\fref{Fig.~\ref}
\newcommand\tref{Tab.~\ref}
\newcommand\sref{Sec.~\ref}

\newcommand\ha{ \rowcolor{orange!0}}
\newcommand\hb{ \rowcolor{orange!15}}
\newcommand\hc{ \rowcolor{orange!40}}

\def\0{{\bf 0}}

\usepackage{amsthm}

\newtheorem{mytheorem}{Theorem}
\newtheorem{assumption}[mytheorem]{Assumption}
\newtheorem{lemma}[mytheorem]{Lemma}
\newtheorem{remk}[mytheorem]{Remark}

\usepackage{tabularx,colortbl}
\usepackage{colortbl}
\usepackage{graphicx}

\usepackage{hyperref}

\newcommand\myshade{85}
\colorlet{mylinkcolor}{violet}
\colorlet{mycitecolor}{YellowOrange}
\colorlet{myurlcolor}{Aquamarine}

\hypersetup{
  linkcolor  = mylinkcolor!\myshade!black,
  citecolor  = mycitecolor!\myshade!black,
  urlcolor   = myurlcolor!\myshade!black,
  colorlinks = true,
}

\newcommand{\OURS}{Q-BERT}
\newcommand{\MP}{\xspace{\tiny{MP}}}
\newcommand{\bertbase}{BERT{\tiny{BASE}}}
\makeatletter
\newcommand\footnoteref[1]{\protected@xdef\@thefnmark{\ref{#1}}\@footnotemark}
\newcommand*{\rom}[1]{\expandafter\@slowromancap\romannumeral #1@}
\makeatother

\usepackage{tikz}
\usepackage[maxbibnames=100,style=numeric,natbib=true,backend=bibtex, sortcites=true, giveninits=true]{biblatex}

\bibliography{ref}

\usepackage{fancyhdr}
\begin{document}
\title{\Large Q-BERT: Hessian Based Ultra Low Precision Quantization of BERT}

\author{
\large Sheng Shen\footnote{Equal Contribution}\ , Zhen Dong$^*$, Jiayu Ye$^*$, Linjian Ma\footnote{Work done while interning at Wave Computing.} , Zhewei Yao,\\
Amir Gholami, Michael W. Mahoney, Kurt Keutzer\\
University of California at Berkeley\\
\normalsize \{sheng.s, zhendong, yejiayu, linjian, zheweiy, amirgh, mahoneymw, and keutzer\}@berkeley.edu
}

\maketitle
\begin{abstract}
Transformer based architectures have become de-facto models used for a range
of Natural Language Processing tasks. In particular, the BERT based models
achieved significant accuracy gain for GLUE tasks, CoNLL-03 and SQuAD.
However, BERT based models have a prohibitive memory footprint and latency.
As a result, deploying BERT based models in resource constrained environments
has become a challenging task. 
In this work, we perform an extensive analysis of fine-tuned BERT models using second order Hessian information, and we use our results to propose a novel method for quantizing BERT models to ultra low precision.
In particular, we propose a new group-wise quantization scheme, and we
use a Hessian based mix-precision method to compress the model further.
We extensively test our proposed method on BERT downstream tasks of SST-2, MNLI, CoNLL-03, and SQuAD.
We can achieve comparable performance to baseline with at most $2.3\%$ performance degradation, even with ultra-low precision quantization down to 2 bits, corresponding up to $13\times$ compression of the model parameters, and up to $4\times$ compression of the embedding table as well as activations.
Among all tasks, we observed the highest performance loss for BERT fine-tuned on SQuAD.
By probing into the Hessian based analysis as well as visualization, we show that this is related to the fact that current training/fine-tuning strategy of BERT does
not converge for SQuAD. 

\end{abstract}
\section{Introduction}

Language model pre-training from large unlabeled 
data has become the new driving-power for models such as
BERT, XLNet, and RoBerta~\cite{devlin2018bert,yang2019xlnet,liu2019roberta}. 
Built upon Transformer~\cite{vaswani2017attention},
BERT based~\cite{devlin2018bert} models significantly improve the state of the art performance when 
fine-tuned on various Natural Language Processing (NLP) tasks~\cite{rajpurkar2016SQuAD,wang2018glue}. 
Recently,  many follow-up works
push this line of research even further by increasing the model capacity 
to more than billions of parameters~\cite{radford2019language}. 
Though these models achieve cutting-edge results on various NLP tasks,
the resulting models have high latency, and prohibitive memory footprint
and power consumption for edge inference.
This, in turn, has limited the deployment of these models on
embedded devices like cellphones or smart assistance, which now require cloud connectivity to function.

A promising method to address this challenge is quantization,
which uses low bit precision for parameter storage and
enables low bit hardware operations to speed up inference. The reduced memory footprint and accelerated inference can then enable edge deployment on hardware that supports reduced precision inference such as FPGAs or
domain specific accelerators.
However, for ultra low-bit 
setting, e.g., 4 bits, the generalization performance of the quantized model can significantly degrade, and this
may not be acceptable for a target application.
Historically, in the computer vision area, a large prominent 
line of work tackles this problem, e.g., different quantization 
schemes~\cite{krishnamoorthi2018quantizing,zhang2018lq}, mixed 
precision quantization~\cite{dong2019hawq,wu2018mixed,zhou2018adaptive}, etc. However, there is very limited work done on 
NLP~\cite{xu2018alternating,wang2018hitnet}, particularly on BERT-based models, which are 
actually more in need of model compression and acceleration.

In this paper, we focus on ultra low precision quantization of BERT based
models, with the goal of minimizing performance degradation while maintaining hardware efficiency.
To achieve this, we incorporate a number of novel techniques and propose \OURS.
The contributions of our work include:

\begin{figure*}[t]
\begin{center}
 \subfloat[MNLI $4^\text{th}$ layer ]{\label{hessain_2:box1}\includegraphics[scale=0.2]{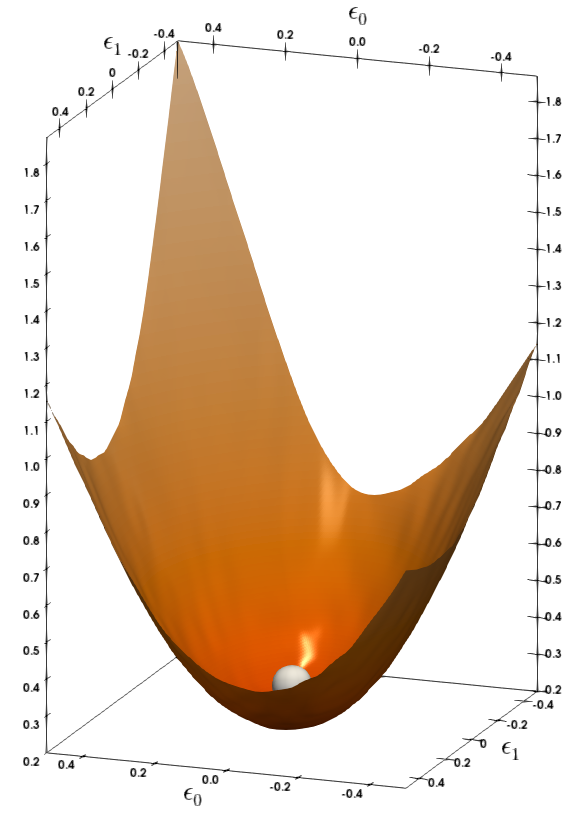}}
  \subfloat[MNLI $10^\text{th}$ layer]{\label{hessain_2:box2}\includegraphics[scale=0.2]{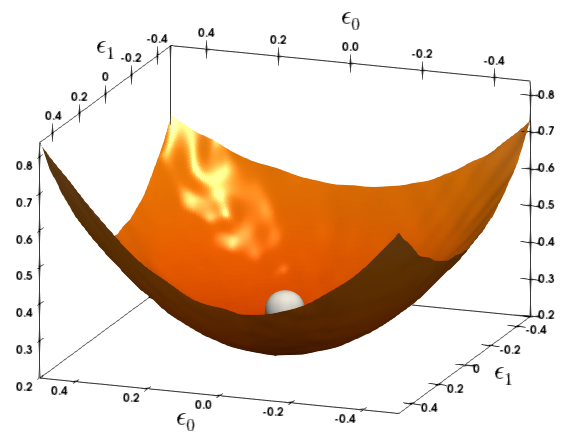}}
  \subfloat[CoNLL-03 $4^\text{th}$ layer ]{\label{hessain_2:box3}\includegraphics[width=.25\linewidth]{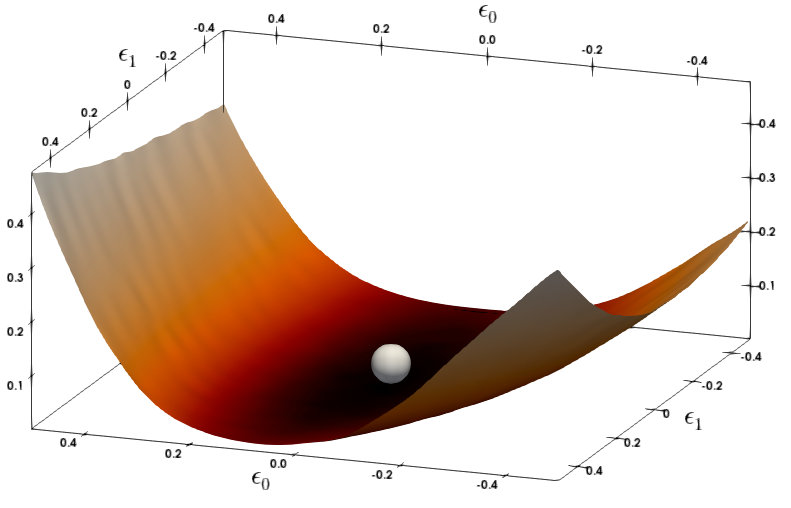}}
  \subfloat[CoNLL-03 $11^\text{th}$ layer ]{\label{hessain_2:box4}\includegraphics[width=.25\linewidth]{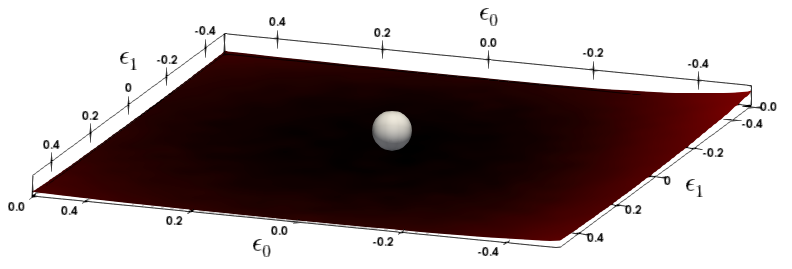}}
\end{center}
 \caption{\footnotesize
 The loss landscape for different layers in MNLI and CoNNL-03 is
 illustrated by perturbing the parameters
 along the first two dominant eigenvectors of the Hessian. 
The silver sphere shows the point in the parameter space to which the BERT model
has converged.
Layers that exhibit flatter curvature can be quantized to lower bit precision.
}
\label{fig:Hessian-loss-landscape-mnli-conll}
\end{figure*}

\begin{itemize}[noitemsep,topsep=0pt,parsep=0pt,partopsep=0pt,leftmargin=*]
    \item We apply mixed-precision quantization on BERT, guided by extensive layer-wise analysis of second order information (i.e., Hessian information). We find that BERT exhibits a drastically different Hessian behaviour, as compared with NN models for computer vision~\cite{yao2018Hessian,dong2019hawq}. 
    Therefore, we propose a sensitivity measurement based on both mean and variance of the top eigenvalues in order to achieve better mixed-precision
    quantization, as opposed to~\cite{dong2019hawq}, which only uses mean~value.
    \item We propose a new quantization scheme, named group-wise quantization, 
    which can alleviate accuracy degradation, without significant increase in hardware complexity.
    Specifically, in group-wise quantization scheme, we partition each
    matrix to different groups, each with its unique quantization range and look up table.
    \item We investigate the bottlenecks in BERT quantization, namely how different factors 
    such as quantization scheme and modules such as embedding, self-attention, and 
    fully-connected layers affect the trade-off between NLP performance and the model compression ratio.
\end{itemize}

    We evaluate \OURS\xspace in four downstream tasks, including Sentiment Classification, 
Natural Language Inference, Named Entity Recognition, and Machine Reading Comprehension. 
\OURS\xspace achieves $13\times$ compression ratio in weights, $4\times$ smaller activation size, and $4\times$ smaller embedding size, within at most 2.3\% accuracy loss.
To the best of our knowledge, this is the first work for BERT quantization 
to ultra low bits with acceptable performance~loss.

\begin{figure*}[!htp]
\begin{center}
  \subfloat[SST-2]{\label{subfig:box1}\includegraphics[width=.45\linewidth]{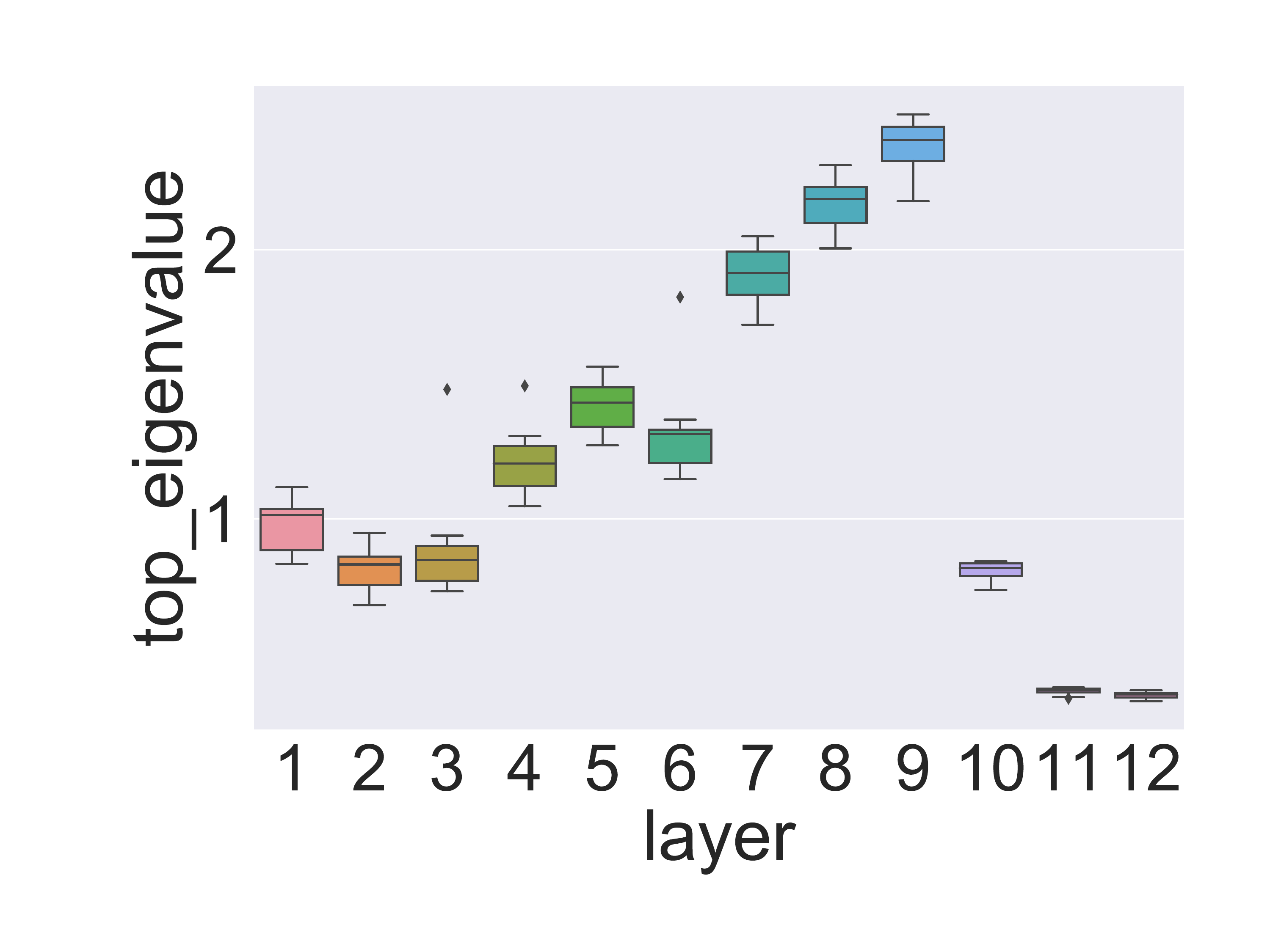}}
  \subfloat[MNLI]{\label{subfig:box2}\includegraphics[width=.45\linewidth]{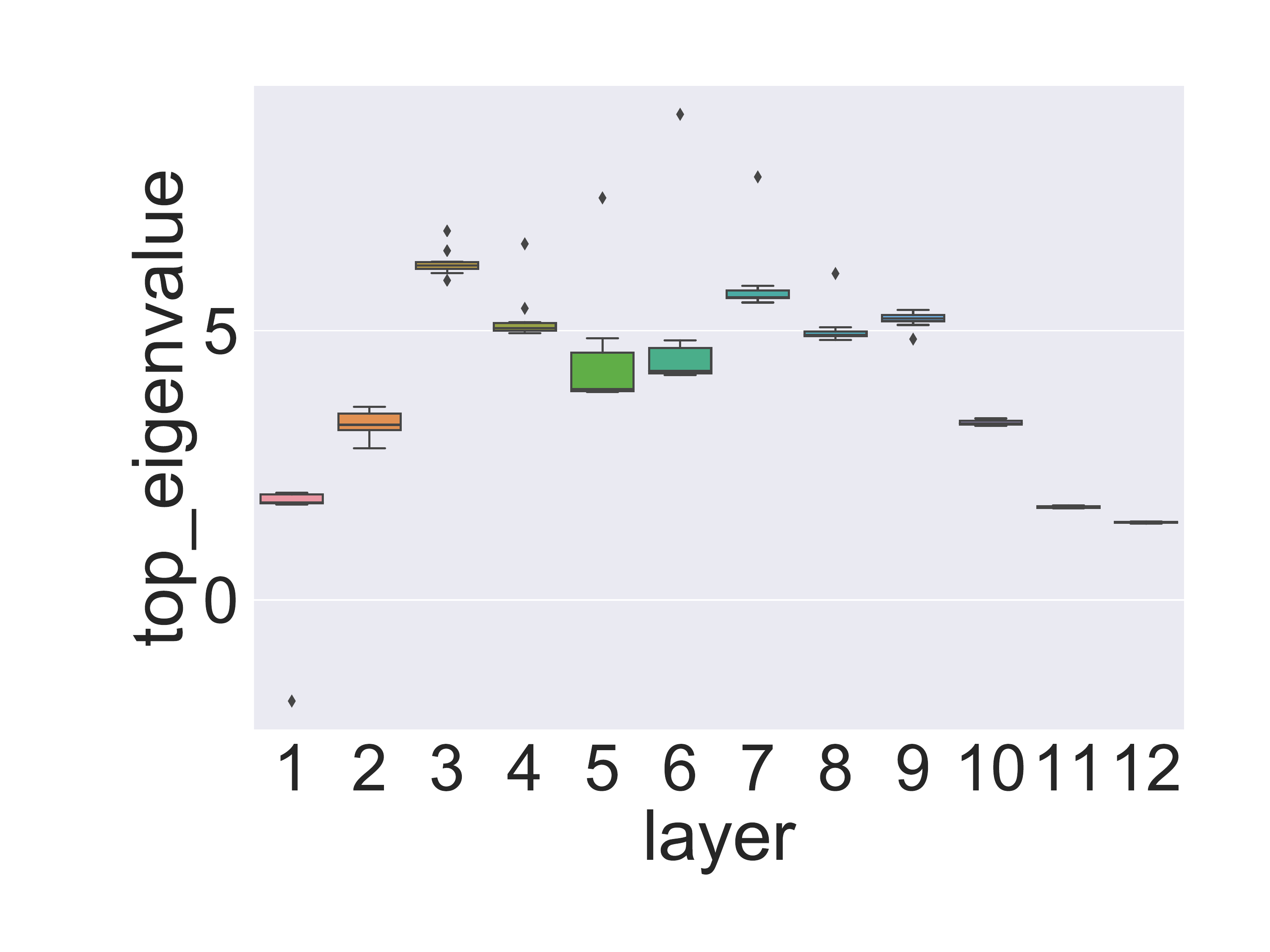}}\\
  \subfloat[CoNLL-03]{\label{subfig:box4}\includegraphics[width=.45\linewidth]{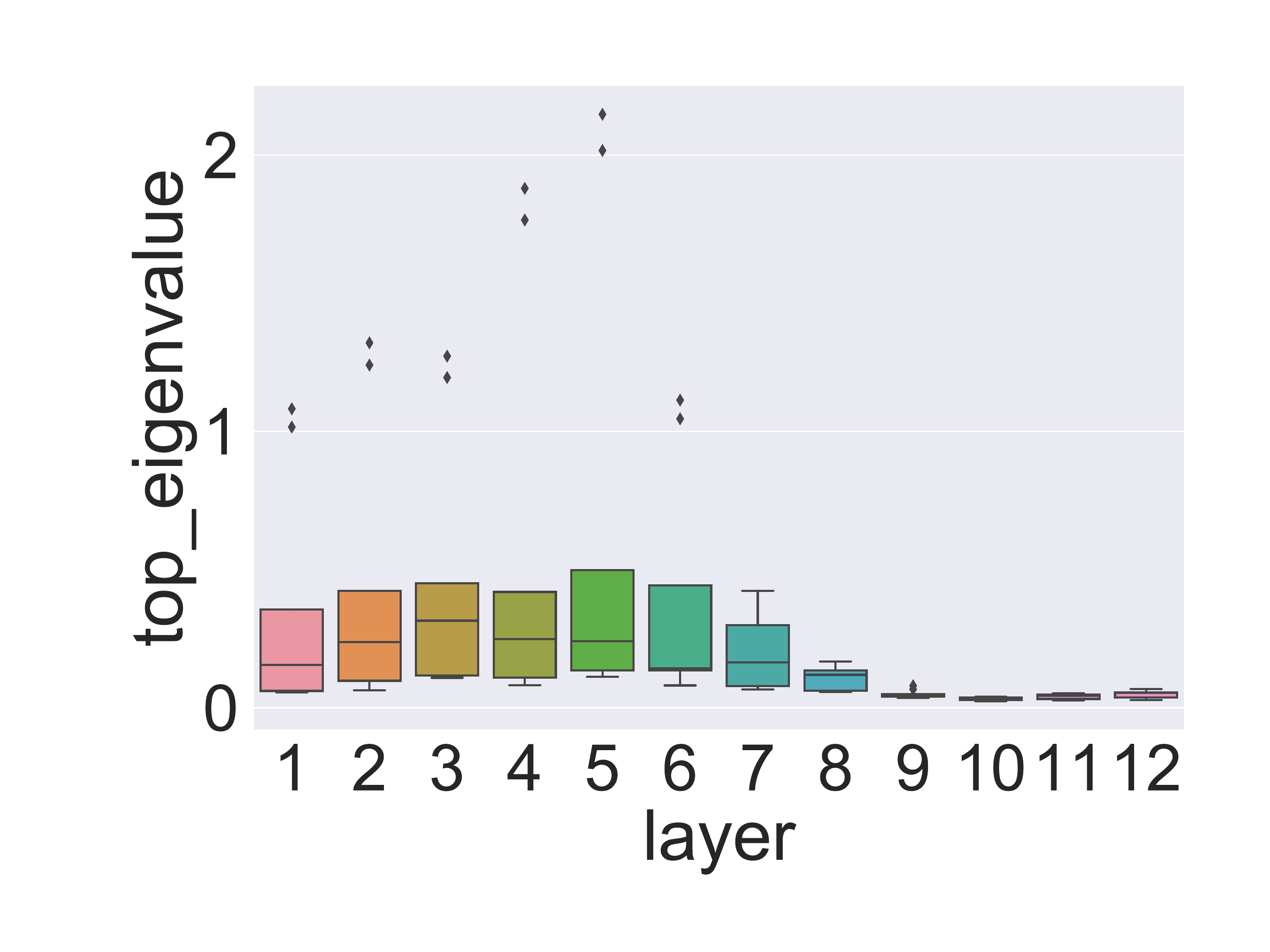}}
  \subfloat[SQuAD]{\label{subfig:box5}\includegraphics[width=.45\linewidth]{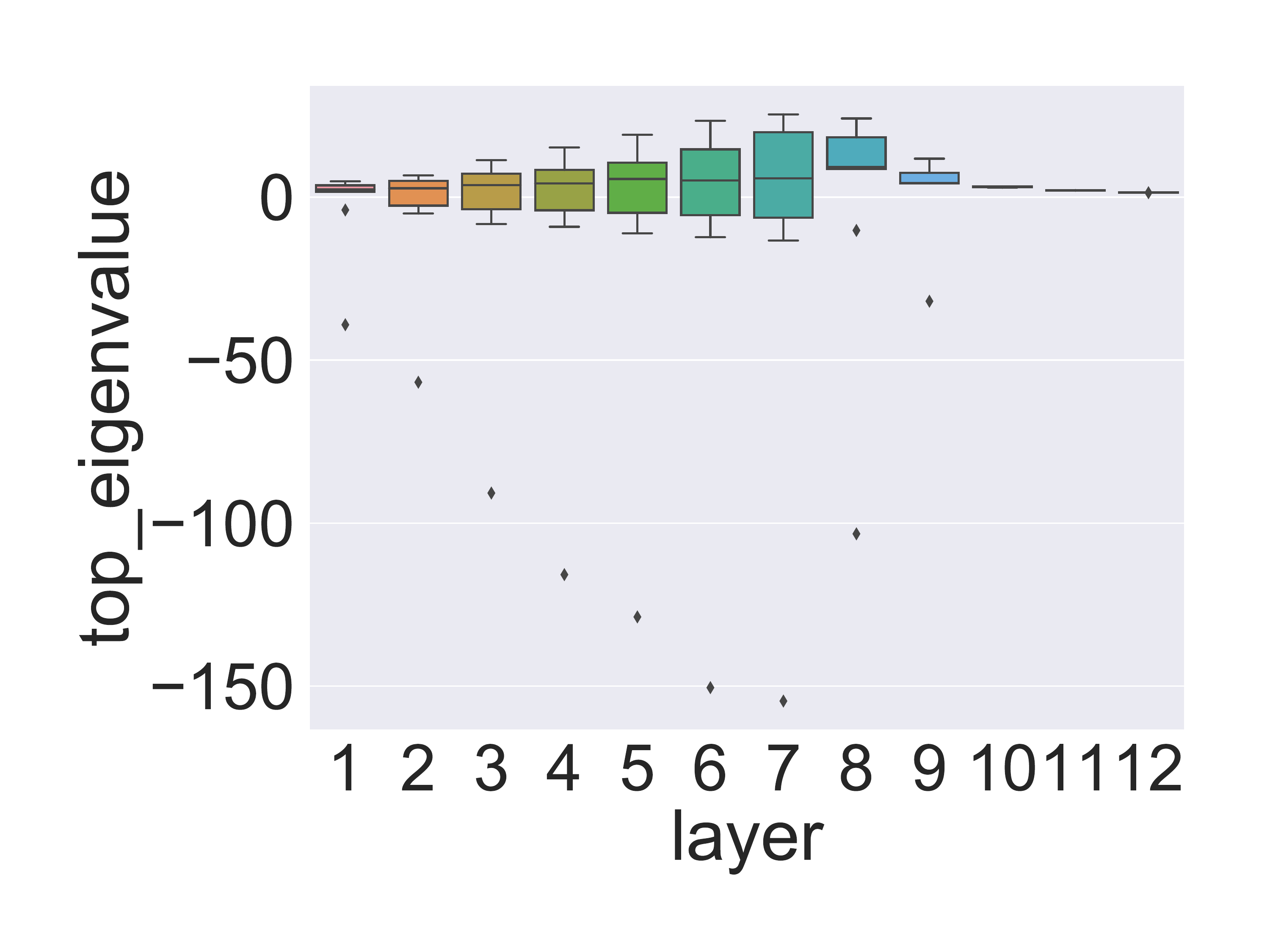}}
\end{center}
 \caption{\footnotesize
 From (a) to (d): Top eigenvalue distributions for different encoder
 layers for SST-2, MNLI, CoNNL-03, SQuAD, respectively. For each task, $10\%$ of the data
 is used to compute the top eigenvalue, and we perform 10 individual runs to plot the top eigenvalue distribution. 
 It can be seen that layers in the middle have higher mean values, and they also tend to have larger variance than the others.
 The last three layers have the smallest variance as well as mean values among all layers.
 }
\label{fig:Hessian-eig-variance}
\end{figure*}

\vspace{10mm}

\section{Related Work}

\paragraph{Model compression}
Model compression is a very active area of research. Efforts in this area could be broadly
categorized as follows:
({\romannumeral1}) new architectures that are compact by design
\cite{iandola2016squeezenet,howard2017mobilenets};
({\romannumeral2}) automated neural architecture search (NAS) with reward function
set as latency or model size
\cite{wang2019haq,wu2019fbnet};
({\romannumeral3}) pruning based methods to reduce model size of existing architectures
\cite{lecun1990optimal,li2016pruning};
({\romannumeral4}) knowledge distillation from a large model to help train a more compact model
\cite{ba2014deep,hinton2015distilling};
({\romannumeral5}) hardware and architecture co-design
\cite{gholami2018squeezenext};
and ({\romannumeral6}) inference quantization
\cite{zhang2018lq,dong2019hawq}.

Here we solely focus on quantization~\cite{courbariaux2015binaryconnect,rastegari2016xnor,li2016ternary,zhou2016dorefa,choi2018pact,Jacob_2018_CVPR,zhang2018lq,dong2019hawq}.
One of the challenges here is that ultra low precision quantization
can lead to significant accuracy degradation. Mixed precision quantization~\cite{wu2018mixed,zhou2018adaptive,wang2019haq} and multi-stage quantization~\cite{zhou2017incremental} have been proposed to solve/alleviate this problem.
However, the challenge with mixed-precision quantization is that the search space is exponentially large. For instance,
if we have three precision options for a specific layer (2, 4 or 8-bits), then the total search 
space of each fine-tuned BERT model~\cite{devlin2018bert} becomes $3^{12} \approx 5.3\times$ $10^{5}$ different precision settings. 
Recently, \cite{dong2019hawq} proposed a 
second-order sensitivity based method to address this issue and achieved state-of-the-art results on 
computer vision tasks. 
Part of our paper builds upon this prior work and extends the results to include other variations of
second order information instead of just the mean value of the Hessian spectrum.

\noindent
\paragraph{Compressed NLP model} 
Notable examples for NLP compression work are
LSTM and GRU-based models for machine translation and language 
model~\cite{xu2018alternating,wang2018hitnet}.
From the recent introduction of Tranformer models, we have observed
a significant increase in NLP model size. This is due to the incorporation
of very large fully connected layers  and attention matrices in Transformers~\cite{vaswani2017attention,devlin2018bert,yang2019xlnet,liu2019roberta,radford2019language}.
Model compression is crucial for deploying these models in resource constrained environments.
Pilot works addressing this are~\cite{michel2019sixteen,bhandare2019efficient}.
From a different angle, \cite{tay2019lightweight,ma2019tensorized} have probed the architectural change of self-attention layer to make the Transformer lightweight.
There have also been attempts to use distillation to reduce large 
pre-trained Transformer models such as BERT~\cite{devlin2018bert} in~\cite{tang2019distilling,sun2019patient}. However, 
significant accuracy loss is observed even for relatively small compression ratio of $4\times$.
Here we show that this compression ratio could be increased up to $13\times$, including $4\times$ reduction
of embedding layer, with much smaller performance degradation.

\section{Methodology}

In this section, we introduce our proposed BERT quantization methods, including the mixed precision 
quantization based on Hessian information, as well as techniques used for the group-wise quantizing scheme.

As in~\cite{devlin2018bert}, a fine-tuned \bertbase~model consists of three parts: embedding; Transformer based encoder layers; and output layer.
Specifically, assuming $x\in X$ is the input word (sentence) and $y\in Y$ is the corresponding label, we have
the loss function $L$ defined as:
\begin{equation*} 
    L(\theta) =\sum_{(x_i,y_i)}
    \text{CE}(\text{softmax}(W_c(W_n(...W_1(W_e(x_i))))), y_i),
\end{equation*}
where CE is the cross entropy function (or other appropriate loss functions), $\theta$ is a combination of $W_e$, $W_1,~W_2,...,W_n$ and $W_c$.
Here, $W_e$ is the embedding table, $W_1,~W_2,...,~W_n$ are the 
encoder layers, and $W_c$ is the output/classifier layer\footnote{Here, we use $W_*$ for both function and its corresponding parameters without confusion.}.

The size of parameters in \mbox{BERT\tiny{BASE}} model is 91MB for embedding, 325MB for encoder and 0.01MB for output. 
We do not quantize the output layer due to its negligible size, and focus on quantizing both the embedding and encoder layers.
As will be discussed in~\sref{sec:quantization_module}, we find
that the embedding layer is more sensitive to quantization than 
the encoder layers.
As a result, we quantize embedding and encoder parameters in different ways. 
The quantization schemes we used are explained in detail in the following sections.

\subsection{Quantization process}\label{sec:quantization_process}
General NN inference is performed in floating point precision for
both weights and activations. Quantization restricts the network weights to
a finite set of values defined as follows:
\begin{equation*}
Q(z) = q_j , \quad\text{for } z \in (t_j , t_{j+1}],    
\end{equation*}
where $Q$ is quantization operator, $z$ is a real valued input tensor (activation or a weight),  and $(t_j , t_{j+1}]$ 
denotes an interval in the real numbers $(j = 0, \ldots , 2^{k}-1)$. Here $k$ is the quantization precision for a specific layer. 

There are multiple choices for quantization function $Q$. Here we use
uniform quantization function, where the range of floating point values in a tensor is equally split~\cite{zhou2016dorefa,hubara2017quantized} 
and then represented by unsigned integers in $\left\{0, \ldots, 2^{k}-1 \right\}$.
It should be noted that a non-uniform quantizer can potentially further increase the accuracy.
However, we solely focus on uniform quantization since it allows more efficient and easier hardware implementation.
To backpropogate gradients through $Q$, which is non-differentiable, we use the Straight-through Estimator (STE)~\cite{bengio2013estimating}.
See Appendix~\ref{sec:appendix_quantizaiton_process} for more details about the forward 
and backward propagation during the entire quantization process.

\subsection{Mixed precision quantization} 

Different encoder layers are attending to different structures~\cite{clark2019does},
and it is expected that they exhibit different sensitivity.
Thus, assigning the same number of bits to all the layers is
sub-optimal. This scenario is more critical if the targeted model size is very small, which requires 
ultra low precision such as 4-bits or 2-bits.  
As a result we explore mixed-precision quantization, where we assign more bits to more
sensitive layers in order to retain performance.

In~\cite{dong2019hawq}, a Hessian AWare Quantization (HAWQ) is developed for mixed-bits assignments.
The main idea is that the parameters in NN layers 
with higher Hessian spectrum (i.e., larger
top eigenvalues) are more sensitive to quantization and require higher precision, as compared to layers with small Hessian spectrum (i.e., smaller top eigenvalues).
However, there exist 7M parameters for each encoder layer in \mbox{BERT\tiny{BASE}}.
Given that the Hessian of each layer is a matrix of size $7M\times7M$, there is a common misconception
that computing second order statistics is infeasible.
However, the Hessian spectrum can be computed by a matrix-free power iteration method~\cite{yao2018Hessian},
and this does not require explicit formation of the operator. To illustrate this, we 
take the first encoder layer as an example. 
Denoting the gradient of the first encoder layer as $g_1$, for a random vector $v$ with the same dimension as $g_1$, we have 
\begin{equation}
    \frac{\partial g_1^Tv}{\partial W_1} = \frac{\partial g_1^T}{\partial W_1}v + g_1^T\frac{\partial v}{\partial W_1} = \frac{\partial g_1^T}{\partial W_1}v = H_1v,
\end{equation}
where $H_1$ is Hessian matrix of the first encoder. Here the second equation comes from the fact that $v$ 
is independent to $W_1$. The top eigenvalue then can be computed by power iteration, as shown
in~Alg.~\ref{alg:power_iteration} in Appendix. We denote $\lambda_i$ as the top eigenvalue of i-th 
encoder layer. 
Using this approach, we show in~\fref{fig:Hessian-eig-variance} the distribution of top Hessian eigenvalue for different layers of \bertbase.
Different layers exhibit different magnitude of eigenvalues even though all layers have exactly 
same structure and size.

\begin{figure}[t]
\begin{center}
  \subfloat[SQuAD $7^\text{th}$ layer ]{\includegraphics[width=.25\linewidth]{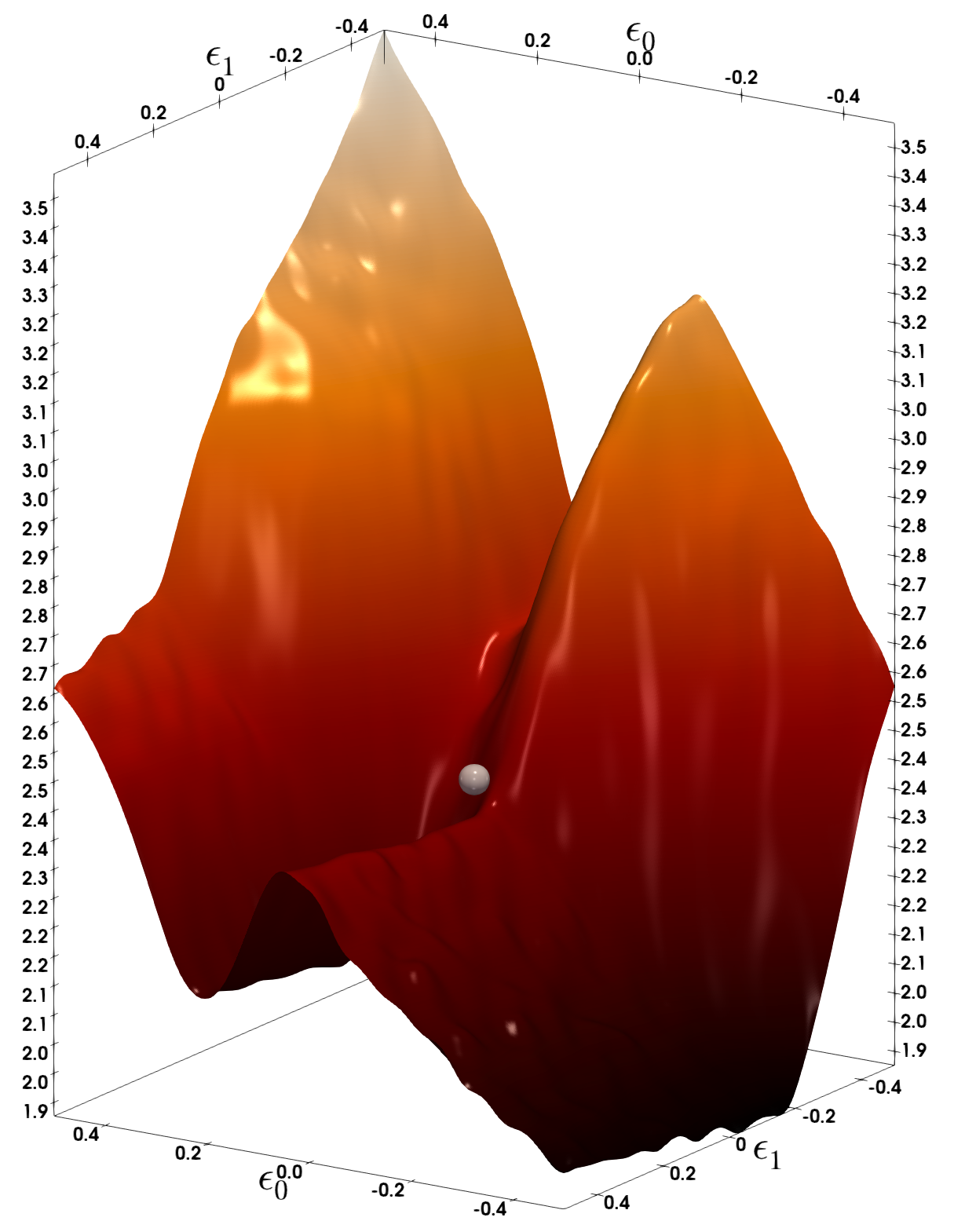}}
  \subfloat[SQuAD $11^\text{th}$ layer]{\label{subfig:Hessian-loss-landscape-squad-11}\includegraphics[width=.25\linewidth]{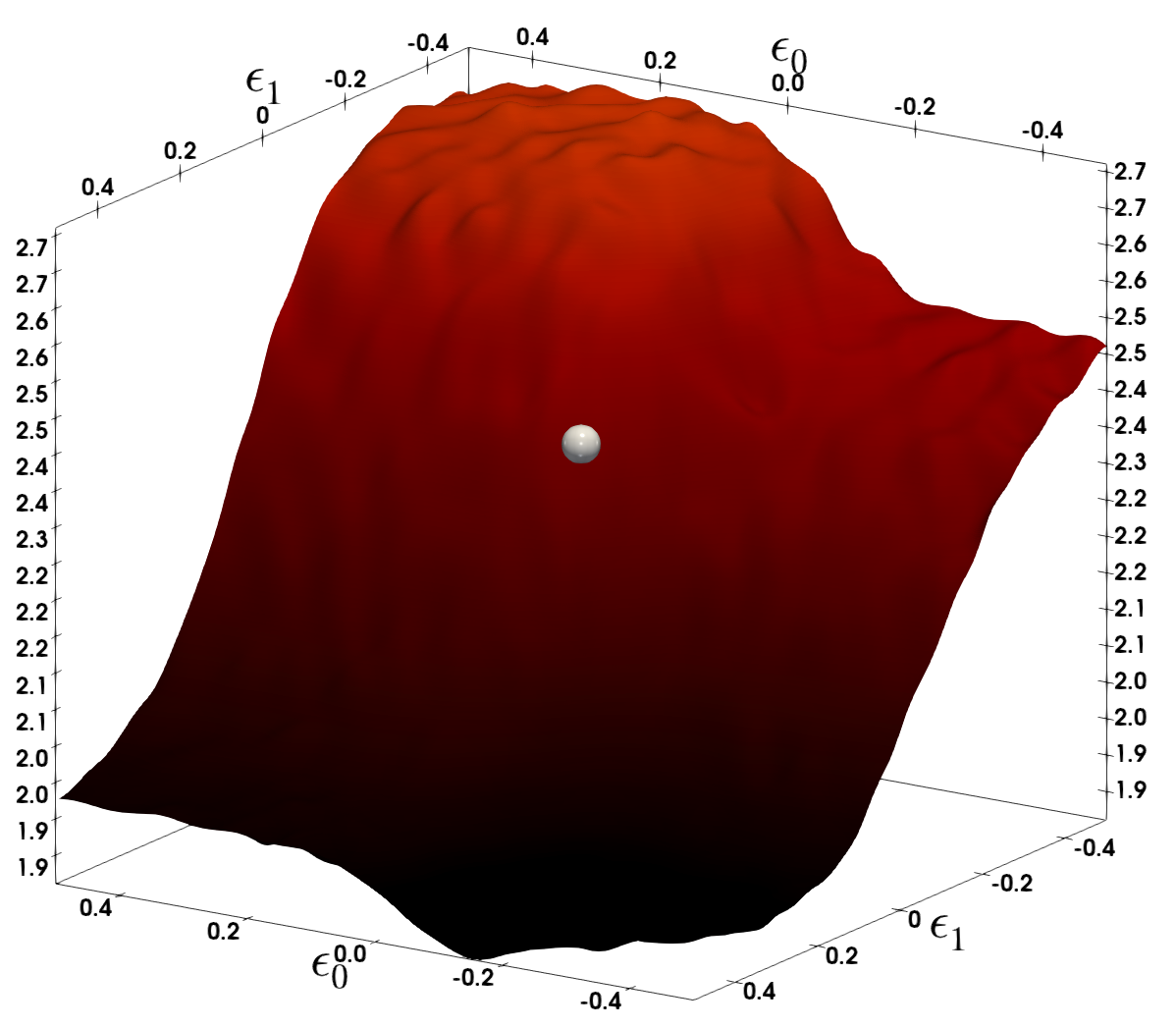}}
\end{center}
 \caption{\footnotesize
  The loss landscape for different layers in SQuAD is
 illustrated by perturbing the parameters
 along the first two dominant eigenvectors of the Hessian. 
The silver sphere shows the point in the parameter space to which the BERT model
has converged.
 Note that the stopping point of SQuAD has 
 negative eigenvalues for both layers. 
 This could be the reason we observed relatively larger performance drop in SQuAD after quantization; see~\tref{tab:squad}.
 }
\label{fig:Hessian-loss-landscape-squad}
\end{figure}

The above Hessian based approach was used in~\cite{dong2019hawq},
where top eigenvalues are computed and averaged for different
training data. More aggressive quantization is performed for layers with
smaller top eigenvalue, which corresponds to flatter loss landscape
as in~\fref{fig:Hessian-loss-landscape-mnli-conll}.
However, we find that assigning bits based only on the average top eigenvalues
is infeasible for many NLP tasks. 
As shown in~\fref{fig:Hessian-eig-variance}, top eigenvalues of Hessian for some layers exhibit very high variance with respect to different portion of the input dataset.
As an example, the variance of the $7^\text{th}$ layer for SQuAD stays larger than 61.6 while the mean of that layer is around 1.0,
even though each data point corresponds to 10\% of the entire dataset (which is 9K samples). 
To address this, we use the 
following metric instead of just using mean value,
\begin{equation}\label{eqn:sensitivity}
\Omega_i\triangleq|\text{mean}(\lambda_i)| + \text{std}(\lambda_i),
\end{equation}
where $\lambda_i$ 
is the distribution of the top eigenvalues of $H_i$, calculated with 10\% of training dataset.%
\footnote{Without confusion, we use $\lambda_i$ for both single top eigenvalue and its distribution with respect to 10\% of the data.}
After $\Omega_i$ is computed, we sort them in descending order, and we use it as a metric to relatively determine the quantization precision. 
We then perform quantization-aware fine-tuning based on the selected precision setting.

An important technical point we need to emphasize is that our method expects that before performing 
quantization the trained model has converged to a local minima. That is, the practitioners
who trained BERT and performed its fine-tuning for downstream tasks should have chosen the hyper-parameters and
number of iterations such that a local minima has been reached. The necessary optimality conditions are zero
gradient, and positive curvature (i.e., positive Hessian eigenvalue).
In our analysis, we observed that for the three tasks of MNLI, CoNLL-03, and SST-2 the top Hessian eigenvalue is indeed positive (see \fref{fig:Hessian-loss-landscape-mnli-conll}, and~\fref{fig:Hessian-loss-landscape-sst2} in Appendix).
However, we find that the BERT model fine-tuned for SQuAD has actually \emph{not} converged
to a local minima, as evident in the Hessian eigenvalues shown in~\fref{fig:Hessian-eig-variance}(d), 
where we observe very large negative eigenvalues. Directly visualizing the loss landscape also shows this very clearly as in~\fref{fig:Hessian-loss-landscape-squad}.
Because of this, our expectation is that performing quantization on SQuAD would lead to higher performance degradation as compared to other tasks,
and this is indeed the case as will be discussed~next.

\begin{figure}[t]
\begin{center}
\centering
    \subfloat[\scriptsize Layer-wise]{
    \label{fig:gp-lw}\
    \includegraphics[width=0.3\linewidth]{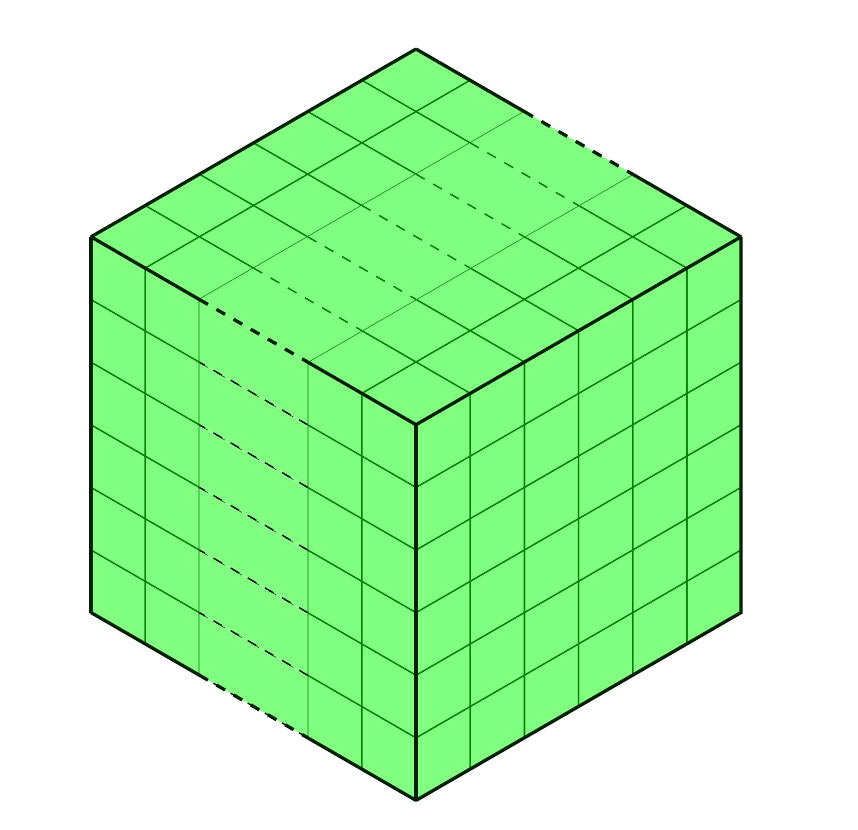}}
    \subfloat[\scriptsize Group-wise ($N_h$ group)]{\label{fig:gp-gpn}\includegraphics[width=0.3\linewidth]{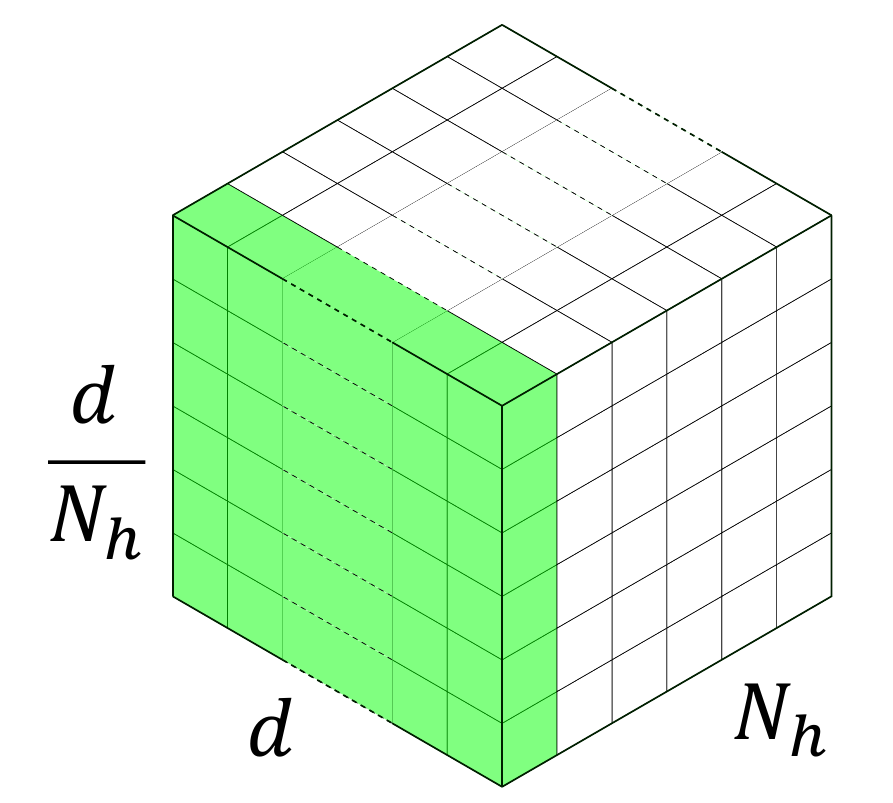}}
    \subfloat[\scriptsize Group-wise ($2N_h$ group)]{\label{fig:gp-gpdd}\includegraphics[width=0.3\linewidth]{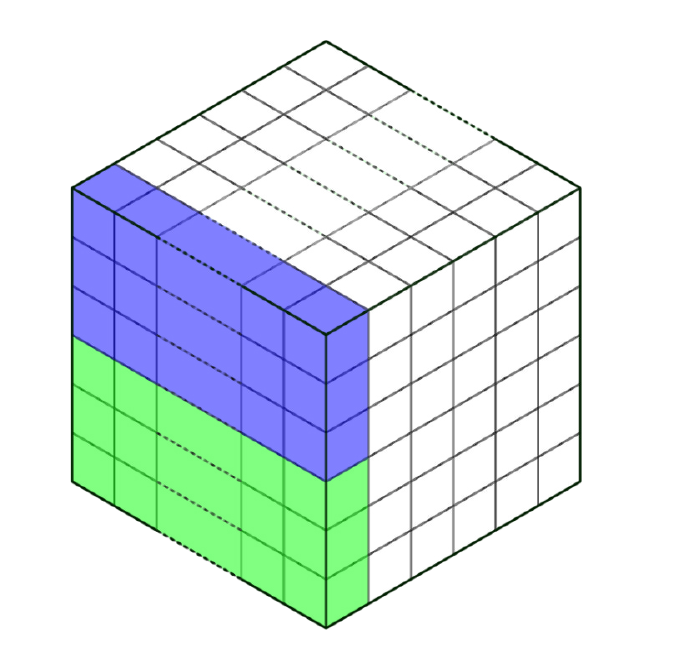}}
\end{center}
 \caption{\footnotesize
 The overview of Group-wise Quantization Method. We illustrate this with value matrices of a multi-head self attention layer. 
    Here $N_h$(number of heads) value matrices $W_v$ are concatenated together, which results in a 3-d tensor.
    The same color denotes the same group with a shared quantization range.
    As shown in (a), for layer-wise quantization, the entire 3-d tensor will be 
    quantized from a universal quantization range into discrete unsigned integers.
    A special case of group-wise quantization in (b) is that we treat each dense matrix as a group, 
    and every matrix can have its own quantization range.
    We show a more general case in (c), where we partition each dense matrix w.r.t. output neuron and bucket every continuous $\frac{d}{2N_h}$ output neurons as a group. 
    }
 \label{fig:gp-overview}
\end{figure}


\subsection{Group-wise Quantization}
Assume that the input sequence has $n$ words and each word has a $d$-dim embedding vector ($d=768$ for \mbox{BERT\tiny{BASE}}), i.e., $x=(x(1), \ldots, x(n))^T\in \mathbb{R}^{n\times d}$.
In Transformer encoder, each self-attention head has 4 dense matrix, i.e., 
$W_k, W_q, W_v\in\mathbb{R}^{\frac{d}{N_h}\times d}, W_o\in\mathbb{R}^{d\times \frac{d}{N_h}}$, where $N_h$ is the number of attention heads. 
Here $W_k$, $W_q$, $W_v$ and $W_o$ stand for key, query, value and output weight matrix.
Each self-attention head computes the weighted sum as
\begin{equation*}
\begin{split}
    \text{Att}({x}, {x(j)})&=W_o\sum_{i=1}^n \text{softmax}\left(\frac{{x(j)}^TW_q^TW_kx(i)}{\sqrt{d}}\right) W_vx(i).
\end{split}
\end{equation*}
Through this reparametrization, the multi-head self-attention (MHSA) will add these features into the final output, that is we will have $\sum_{i=1}^{N_h} \text{Att}_i({x}, {x(j)})$.
Directly quantizing each 4 matrices in MHSA as an entirety with the 
same quantization range can significantly degrade the accuracy, since there are 
more than 2M parameters in total, which corresponds to $4\times12\times64=3072$ neurons,
and the weights corresponding to each neuron may lie in different range of real numbers.
Channel-wise quantization can be used to alleviate this problem in convolutional neural networks, 
where each convolutional kernel can be treated as a single output channel 
and have its own quantization range. 
However, this cannot be directly applied for dense matrices, since each dense matrix itself is a single kernel.
Therefore, we propose group-wise quantization for attention-based models. 
We treat the individual matrix $W$ with respect to each head in one dense matrix of MHSA as a group so there will be $12$ groups.
Furthermore, in each group, we bucket sequential output neurons together as sub-groups, e.g., each
6 output neurons as one sub-group so there are $12\times\frac{64}{6}=128$ sub-group in total (the hidden dimension in each head of \bertbase~is $\frac{768}{12}=64$). 
Each sub-group can have its own quantization range. 
An illustration is shown in~\fref{fig:gp-overview} for $W_v$, 
where we concatenate $N_h$ value matrix $W_v$ to be a 3-d tensor. For layer-wise quantization, the entire 3-d tensor will be quantized into the same range of discrete numbers, as shown in~\fref{fig:gp-lw}. 
A special case of group-wise quantization is that we treat each dense matrix as a group, 
and every matrix can have its own quantization range as shown in~\fref{fig:gp-gpn}. 
A more general case in~\fref{fig:gp-gpdd} is that we partition each dense matrix with respect to output neuron, and we bucket every continuous $\frac{d}{2N_h}$ output neurons as a group.
The effect of finer group-wise quantization is further investigated in~\sref{sec:result_groupwise}.

\section{Experiment}
\label{sec:main_result} 
In this section, we describe our experiments on evaluating the proposed \OURS~on four different NLP tasks. Details of the datasets are shown in 
Appendix~\ref{sec:appendix_dataset}. To the best of our knowledge, there is no published work done on 
BERT quantization at this point, so we report Direct quantization (DirectQ), i.e., quantization without 
mixed-precision and group-wise quantization as a baseline.

\vspace{-2mm}
\begin{table}[!htbp]
    \caption{
    \footnotesize
    Quantization results for {BERT\tiny{BASE}} on Natural Language Understanding tasks. 
    Results are obtained with 128 groups in each layer.
    We abbreviate quantization bits used for weights as ``w-bits'', embedding as ``e-bits'', model size in MB as ``Size'', and model size without embedding layer in MB as ``Size-w/o-e''. For simplicity and efficacy, all the models except for Baseline are using 8\text{-bits} activation. Furthermore, we compare \OURS~with direct quantization method (``DirectQ'') without using mixed precision or group-wise quantization. Here ``MP'' refers to mixed-precision quantization.
    }
\setlength\tabcolsep{2.35pt}
\centering
\subfloat[\footnotesize SST-2]{
\centering
\begin{tabular}{lcccccccccccccc}
\toprule
Method      & w-bits    & e-bits    & Acc           & Size      & Size-w/o-e\\
\midrule        
Baseline    & 32        & 32        & 93.00         & 415.4     & 324.5 \\
\midrule            
\OURS       & 8         & 8         & 92.88         & 103.9     & 81.2\\
\midrule                    
DirectQ     & 4         & 8         & 85.67         & 63.4      & 40.6\\
\hc \OURS   & 4         & 8         & \bf{92.66}    & 63.4      & 40.6 \\ 
\midrule                        
DirectQ     & 3         & 8         & 82.86         & 53.2      & 30.5      \\
\hb \OURS   & 3         & 8         & \bf{92.54}    & 53.2      & 30.5      \\
\hc\OURS\MP & 2/4~\MP   & 8         & \bf{92.55}    & 53.2      & 30.5      \\
\midrule                        
DirectQ     & 2         & 8         & 80.62         & 43.1      & 20.4      \\
\hb \OURS   & 2         & 8         & \bf{84.63}    & 43.1      & 20.4      \\ 
\hc\OURS\MP & 2/3~\MP   & 8         & \bf{92.08}    & \bf{48.1} & \bf{25.4} \\
\bottomrule 
\end{tabular}
\label{tab:sst}
}
\vspace{-2mm}
\subfloat[\footnotesize MNLI]{
\centering
\begin{tabular}{lcccccccccccccc}
\toprule
Method          & w-bits    & e-bits    & Acc           & Acc       &   Size    &   Size\\
                &           &           & m             & mm        &           &   w/o-e\\
\midrule
Baseline        & 32        & 32        & 84.00         & 84.40     & 415.4   &   324.5 \\
\midrule 
\OURS           & 8         & 8         & 83.91         & 83.83     & 103.9 & 81.2\\
\midrule 
DirectQ         & 4         & 8         & 76.69         & 77.00     & 63.4 & 40.6\\
\hc \OURS       & 4         & 8         & \bf{83.89}    & \bf{84.17}    & 63.4 & 40.6\\
\midrule 
DirectQ         & 3         & 8         & 70.27         & 70.89     & 53.2 & 30.5\\
\hb \OURS       & 3         & 8         & \bf{83.41}    & \bf{83.83}    & 53.2 & 30.5\\
\hc \OURS\MP    & 2/4~\MP   & 8         & \bf{83.51}    & \bf{83.55}    & 53.2 & 30.5\\
\midrule 
DirectQ         & 2         & 8         & 53.29         & 53.32         & 43.1 & 20.4\\
\hb \OURS       & 2         & 8         & \bf{76.56}    & \bf{77.02}    & 43.1 & 20.4\\
\hc \OURS\MP    & 2/3~\MP   & 8         & \bf{81.75}    & \bf{82.29}    & \bf{46.1} & \bf{23.4}\\
\bottomrule
\end{tabular}
\label{tab:mnli}
}

\vspace{-1mm}
\subfloat[\footnotesize CoNLL-03]{
\centering
\begin{tabular}{lcccccccccccccc}
\toprule
Method          & w-bits    & e-bits    & F$_1$         & Size      & Size-w/o-e    \\
\midrule                    
Baseline        & 32        & 32        & 95.00         & 410.9     & 324.5         \\
\midrule                                
\OURS           & 8         & 8         & 94.79         & 102.8     & 81.2          \\
\midrule                                
DirectQ         & 4         & 8         & 89.86         & 62.2      & 40.6          \\
\hc \OURS       & 4         & 8         & \bf{94.90}    & 62.2      & 40.6          \\
\midrule                                    
DirectQ         & 3         & 8         & 84.92         & 52.1      & 30.5          \\
\hb \OURS       & 3         & 8         & \bf{94.78}    & 52.1      & 30.5          \\
\hc \OURS\MP        & 2/4~\MP     & 8         & \bf{94.55}    & 52.1      & 30.5          \\ 
\midrule                                    
DirectQ         & 2         & 8         & 54.50         & 42.0      & 20.4          \\
\hb \OURS       & 2         & 8         & \bf{91.06}    & 42.0      & 20.4          \\
\hc \OURS\MP    & 2/3~\MP     & 8         & \bf{94.37}    & \bf{45.0} & \bf{23.4}     \\ 
\bottomrule 
\end{tabular}
\label{tab:ner}
}
\vspace{-1mm}
\subfloat[\footnotesize SQuAD]{
\centering
\small
\centering
\begin{tabular}{lccccccccccccccc}
\toprule
Method          & w-bits    & e-bits    & EM            & F$_1$         & Size      & Size-w/o-e\\
\midrule                    
Baseline        & 32        & 32        & 81.54         & 88.69         & 415.4     & 324.5     \\
\midrule                                
\OURS           & 8         & 8         & 81.07         & 88.47         & 103.9     & 81.2      \\
\midrule                            
DirectQ         & 4         & 8         & 66.05         & 77.10         & 63.4      & 40.6      \\
\hc \OURS       & 4         & 8         & \bf{80.95}    & \bf{88.36}    & 63.4      & 40.6      \\ 
\midrule                                
DirectQ         & 3         & 8         & 46.77         & 59.83         & 53.2      & 30.5      \\
\hb \OURS       & 3         & 8         & \bf{79.96}    & \bf{87.66}    & 53.2      & 30.5      \\
\hc \OURS\MP        & 2/4~\MP     & 8   & \bf{79.85}    & \bf{87.49}    & 53.2      & 30.5      \\
\midrule                                    
DirectQ         & 2         & 8         & 4.77          & 10.32         & 43.1      & 20.4      \\
\hb \OURS       & 2         & 8         & \bf{69.68}    & \bf{79.60}    & 43.1      & 20.4      \\ 
\hc \OURS\MP    & 2/3~\MP     & 8         & \bf{79.29}    & \bf{86.95}    & \bf{48.1} & \bf{25.4} \\
\bottomrule 
\end{tabular}
\label{tab:squad}
}
\label{tab:main}
\end{table}


\subsection{Main Results}
\label{sec: main}

We present results of \OURS~on the development set of the four tasks of SST-2, MNLI, CoNLL-03, and SQuAD, as
summarized in~\tref{tab:main}.
As one can see, \OURS~performs significantly better compared to the DirectQ method across all 
four tasks in each bit setting. 
The gap becomes more obvious for ultra low bit setting. 
As an example, in 4-bits setting, Direct quantization (DirectQ) of SQuAD results in 
11.5\% performance degradation as compared to \bertbase.
However, for the same 4-bits setting, \OURS~only exhibits 0.5\% performance degradation.
Moreover, under 3-bits setting, 
the gap between \OURS~and DirectQ increases even further to 9.68-27.83\% for various tasks.

In order to push further the precision setting to lower bits, we investigate the 
mixed-precision \OURS~(\mbox{\OURS\MP}). As can be seen, \OURS~with uniform 2-bits 
setting has very poor performance across all four tasks, though the memory is
reduced by 20\% against 3-bits setting. 
The reason behind this is the discrepancy that not all the layers have
the same sensitivity to quantization as evident from loss landscape visualizations; see~\fref{fig:Hessian-loss-landscape-mnli-conll} (and~\fref{fig:Hessian-loss-landscape-sst2} in Appendix). 
Intuitively, for more sensitive layers, higher bit precision needs to be set, while 
for layers that are less sensitive, 2-bits setting is already sufficient.
To set mixed precision to each encoder layer of \bertbase, we measure the sensitivity based on~\eref{eqn:sensitivity},
which captures both mean and variance of the top eigenvalue of the Hessian shown in~\fref{fig:Hessian-eig-variance}. 
Note that all experiments in~\fref{fig:Hessian-eig-variance} are based 
on 10 runs and each run uses 10\% of the entire training dataset. 
We can obverse that for most of the lower encoder layers (layer 1-8), 
the variance is pretty large compared to the last three layers.
We generally observe that the middle part (layer 4-8) has the
largest $\text{mean}(\lambda_i)$. 
Beyond the relatively smaller mean, the last three layers also have much smaller variance, which indicates the insensitivity of these layers. 
Therefore, higher bits will only be assigned for middle layers according to~\eref{eqn:sensitivity} for \mbox{\OURS~2/3 \MP}.%
\footnote{Exact detailed bits setting is included in the Appendix~\ref{sec:ab-hessian-bits}}
In this way, with only additional 5MB memory storage, 
2/3-bits \mbox{\OURS\MP} is able to retain the performance drop 
within 2.3\% for MNLI, SQuAD and 1.1\% for SST-2, CoNLL-03, with up to $13\times$ 
compression ratio in weights. Note that this is up to 7\% better than \OURS~with 
uniform 2 bits.

One consideration for quantization is that 3-bit quantized
execution is typically not supported in hardware. It is however possible
to load 3-bit quantized values and cast them to higher bit precision such
as 4 or 8 bits in the execution units. This would still have the benefit
of reduced memory volume to/from DRAM. It is also possible to avoid using
3 bits and instead use a mixture of 2 and 4 bits as shown in~\tref{tab:main}.
For example, SST-2 \OURS\MP~with mixed 2/4-bit precision weights
has the same model size as the 3 bit quantization in 53.2MB and achieves
similar accuracy. We observe a similar trend for other tasks as well.

One important observation is that we found SQuAD to be harder to quantize
as compared to other tasks; see~\tref{tab:squad}.
For example, 2-bits DirectQ results in only 10\% F$_1$ score.
Even \OURS~has larger performance drop as compared to other tasks in~\tref{tab:main}.
We studied this phenomenon further through Hessian analysis. 
In~\fref{fig:Hessian-eig-variance}, among all the tasks, it can be clearly seen that SQuAD not only has much larger eigenvalue variance, but it has very large negative eigenvalues. 
In fact this shows that the existing BERT model for SQuAD has not reached a local
minima.
This is further illustrated in the 3-d loss landscape of all four tasks 
in~\fref{fig:Hessian-loss-landscape-mnli-conll} and~\fref{fig:Hessian-loss-landscape-squad}   (and~\fref{fig:Hessian-loss-landscape-sst2} in Appendix). 
It can be clearly seen that for the other three tasks, the stopping point 
is at a quadratic bowl (at least in the first two dominant eigenvalue directions of the Hessian). However, compared to the others, SQuAD has a totally different 
structure to its loss landscape. As shown in~\fref{fig:Hessian-loss-landscape-squad}, the stopping points of different layers on SQuAD have negative curvature directions, 
which means they have not converged to a local minima yet. 
This could well explain why the quantization of SQuAD results in more accuracy drop. 
Our initial attempts to address this by changing training hyper-parameters were
not successful. We found that the BERT model quickly overfits the training data.
However, we emphasize that fixing BERT model training itself is outside the scope
of this paper and not possible with academic computational resources.

\subsection{Effects of group-wise quantization}\label{sec:result_groupwise}
We measure the performance gains with different group numbers
in~\tref{tab:gp}. 
We can observe from the table that performing layer-wise quantization (shown in~\fref{fig:gp-lw})
is 
sub-optimal for all three tasks (the performance drop is around 7\% even with 4-bits weights).
However, the performance significantly increases as we increase the number of groups.
For example, in the case with 12 groups, the performance degradation is less than 1\% for all
the tasks.
Further increasing the group number from 12 to 128 increases the accuracy by at least 0.3\% accuracy.
However, increasing the group number further from 128 to 768 can only increase the performance within 0.1\%. 
This shows that the performance gain saturates with group number 128, 
and we should note that the saturation group number can potentially be even smaller, such as 12, 
with less aggressive quantization.
It is also preferable not to have very large value for the number of group since
it increases the number of Look-up Tables (LUTs) necessary for each
matrix multiplication.  This can adversely affect hardware performance, and
based on our results there are diminishing returns in terms of accuracy.
To ensure continuity, we used 128 groups for
both \OURS~and~\mbox{\OURS\MP} in all our experiments in~\sref{sec: main}.

\begin{table}[!htbp]
    \caption{\footnotesize
    Effects of group-wise quantization for \OURS~on three tasks. The 
    quantization bits were set to be 4 for weights, 8 for embeddings and 8 for 
    activations on all the tasks. From top to down, we increase the number of 
    groups. In order to balance the accuracy and hardware efficiency, we set 128 
    groups for other experiments.}

\setlength\tabcolsep{2.35pt}

\centering
\begin{tabular}{lccccc}
\toprule
\# Group                & SST-2     & MNLI-m/mm     & CoNLL-03  \\
\midrule
Baseline                & 93.00     & 84.00/84.40   & 95.00     \\
\midrule
 \ha 1                  & 85.67     & 76.69/77.00   & 89.86    \\
\hc 12                  & 92.31     & 83.30/83.55   & 94.42     \\
\ha 128                 & 92.66     & 83.89/84.17   & 94.90     \\
\hc 768 \footnotemark{} & 92.78     & 84.00/84.20   & 94.99     \\
\bottomrule 
\end{tabular}
\label{tab:gp}
\end{table}

\footnotetext{Here we treat each output neuron as a single group.}

\section{Discussion}
In this Section, we further investigate the quantization effects on different modules,
such as different embedding layers (e.g., word and position embeddings), and we perform 
qualitative analysis using attention distribution.  This illustrates
that \OURS~better captures the behaviour of the original model
as compared to DirectQ in all cases. 

\subsection{Quantization effects on different modules} \label{sec:quantization_module}
Here we investigate the quantization effects with respect to different modules of BERT model
(multi-head self-attention versus feed-forward network, and different embedding layers, 
i.e., word embedding versus position embedding). 

Generally speaking, we find that embedding layer is more sensitive than weights for quantization. 
This is illustrated in~\tref{tab:emb}, where we use 4-bits layerwise quantization for embedding,
which results in an unacceptable performance drop up to 10\% for SST-2, MNLI, CoNLL-03 and even more than 20\% for SQuAD. 
This is despite the fact that we used 8/8-bits for weights/activations.
On the contrary, encoder layers consume around 79\% total parameters ($4\times$ embedding parameter size), while quantizing them to 4-bits in~\tref{tab:main} leads to less performance loss.

Furthermore, we find that position embedding is very sensitive
to quantization. For instance, quantizing position embedding to 4 bits results in generally 2\% additional performance degradation than quantizing word embedding, even though the position embedding only accounts for less than 5\% of the entire embedding.
This indicates the importance of positional information in Natural Language Understanding tasks. 
Given position embedding only accounts for a small portion of model size, 
we can do mixed-precision quantization for embedding to further push down the model size boundary with a tolerable accuracy drop, as shown in Appendix~\ref{sec:emb-mix-quantize}.

\begin{table}[!htbp]
\setlength\tabcolsep{2.2pt}    
    \caption{\footnotesize
    Quantization effect to different modules. We abbreviate the quantization bits used for word embedding as ``ew-bits'', position embedding as ``ep-bits'', multi-head attention layer as ``s-bits'' and fully-connected layer as ``f-bits''. In (a), we set weight and activation bits as 8. In (b), we set embedding and activation bits as 8. }

\centering
\small
\subfloat[\footnotesize quantization effect on embedding]{
\hspace*{-0.2cm}\begin{tabular}{lcccccccccc}
\toprule
Method      & ew-bits   & ep-bits   & SST-2     & MNLI-m/mm     & CoNLL-03  & SQuAD \\
\midrule
Baseline    & 32        & 32        & 93.00     & 84.00/84.40   & 95.00     & 88.69 \\
\midrule
\hc \OURS   & 8         & 8         & 92.88     & 83.83/83.91   & 94.79     & 88.47 \\
 \OURS      & 4         & 8         & 91.74     & 82.91/83.67   & 94.44     & 87.55 \\
\hc \OURS   & 8         & 4         & 89.11     & 82.84/82.25   & 93.86     & 72.38 \\ 
\OURS       & 4         & 4         & 85.55     & 78.08/78.96   & 84.32     & 61.70 \\ 
\bottomrule 
\end{tabular}
\label{tab:emb}
}

\subfloat[\footnotesize  quantization of multi-head attention versus fully-connected layer]{
\centering
\hspace*{-0.2cm}\begin{tabular}{lcccccccccc}
\toprule
Method          & s-bits    & f-bits    & SST-2 & MNLI-m/mm       & CoNLL-03  & SQuAD \\
\midrule
Baseline        & 32        & 32        & 93.00 & 84.00/84.40     & 95.00     & 88.69 \\
\midrule
\hc \OURS\MP    & 1/2\MP      & 2/3\MP      & 89.56 & 73.66/74.52     & 91.74     & 75.81 \\ 
\OURS\MP        & 2/3\MP      & 1/2\MP      & 85.89 & 70.89/71.17     & 87.55     & 68.71\\
\hc \OURS\MP    & 2/3\MP      & 2/3\MP      & 92.08 & 81.75/82.29     & 94.37     & 86.95\\
\bottomrule 
\end{tabular}
\label{tab:sf}
}
\label{tab:qt_md}
\end{table}

To study the quantization effects on self-attention layers and fully-connected networks,
we conducted extensive experiments under different bits settings for the encoder layers. 
The results are shown in~\tref{tab:sf}. 
Specifically, we adopt the \OURS\MP~setting in~\tref{tab:main}, with
a mixture of 2 and 3 bits for encoder weights. To test the robustness of the
two modules inside each encoder layer, we further reduce one more bit in the
corresponding modules and denote the resulting precision setting~1/2\MP. From~\tref{tab:sf}, we can conclude that generally self-attention layer is more robust to quantization than the fully-connected network, since 1/2\MP~self-attention results in 
about 7\% performance drop while 1/2\MP~fully-connected will worsen this to 11\%.

\vspace{-2mm}
\subsection{Qualitative Analysis}
We use attention information to conduct qualitative analysis to 
analyze the  difference between \OURS~and DirectQ.

\begin{figure}[]
\begin{center}
    \subfloat[SST-2]{\includegraphics[width=0.45\linewidth]{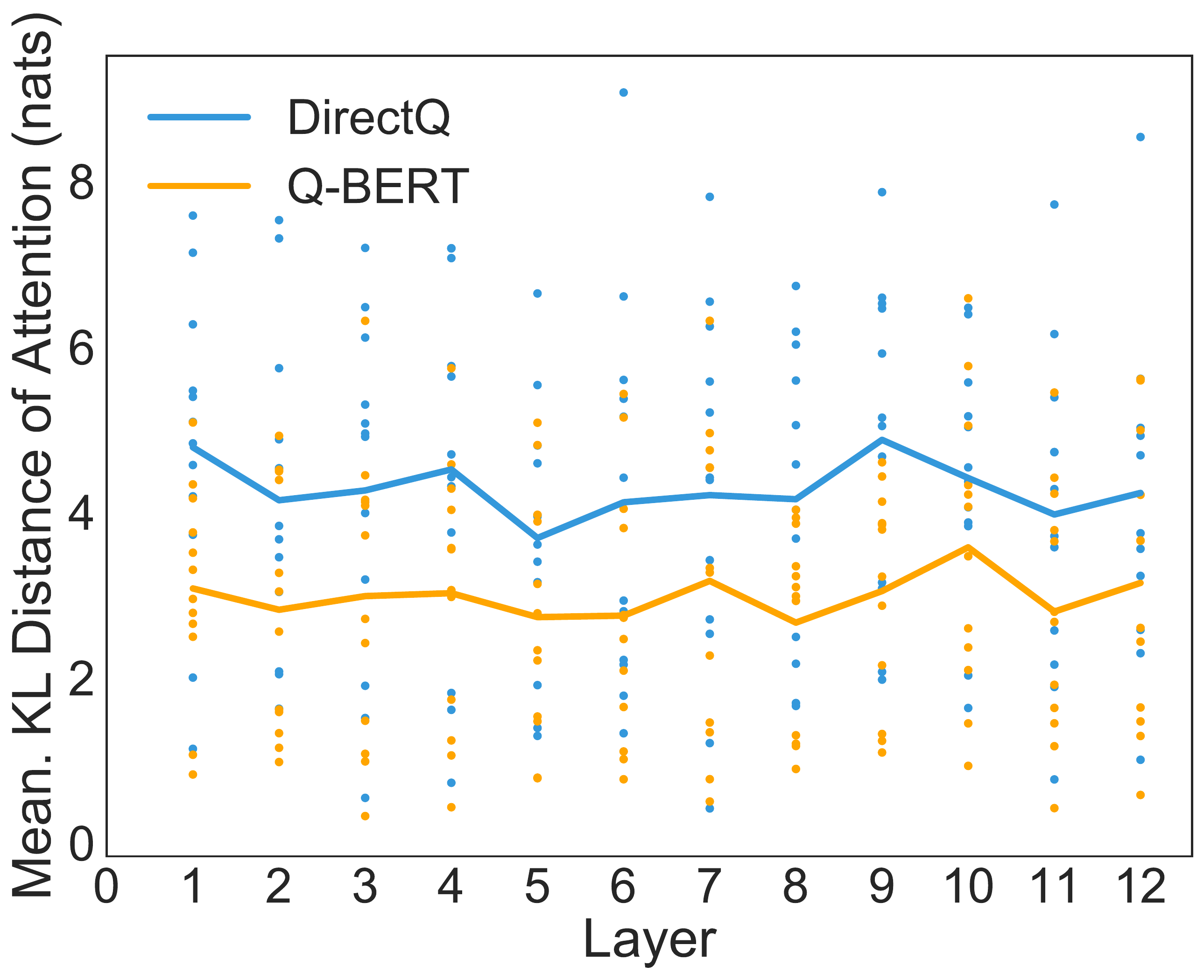}}
    \subfloat[MNLI]{\includegraphics[width=0.45\linewidth]{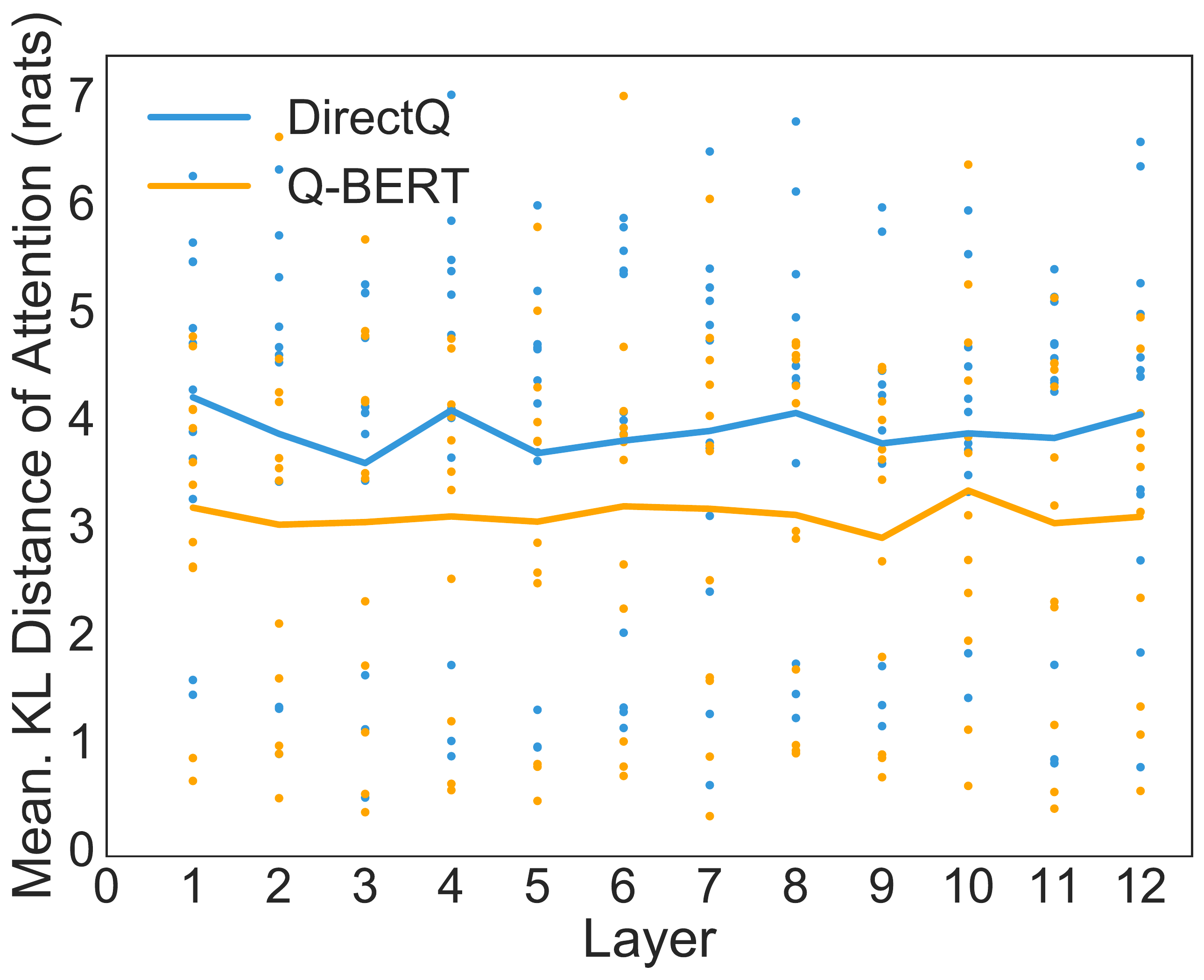}}

    \subfloat[CoNLL-03]{\includegraphics[width=0.45\linewidth]{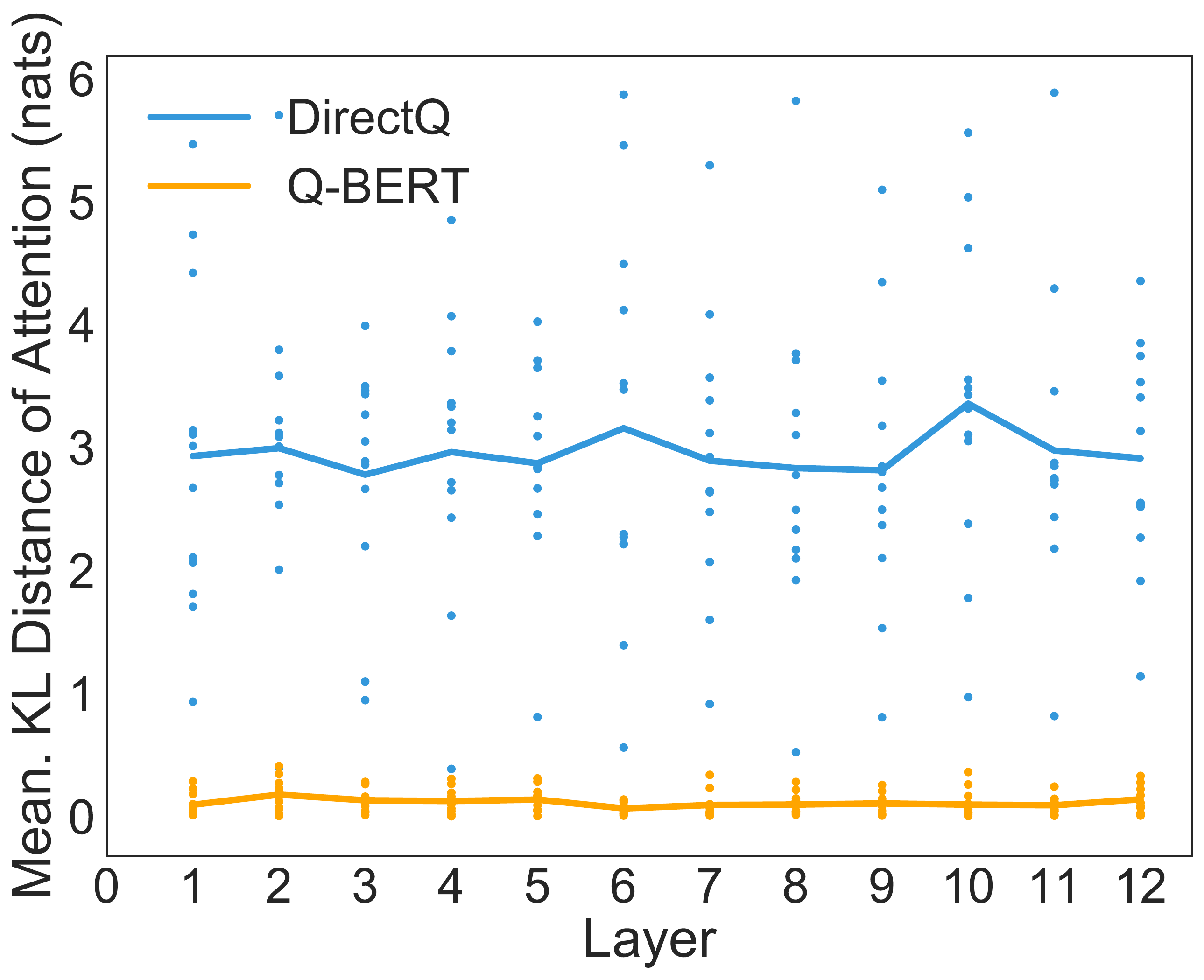}}
    \subfloat[SQuAD]{\includegraphics[width=0.45\linewidth]{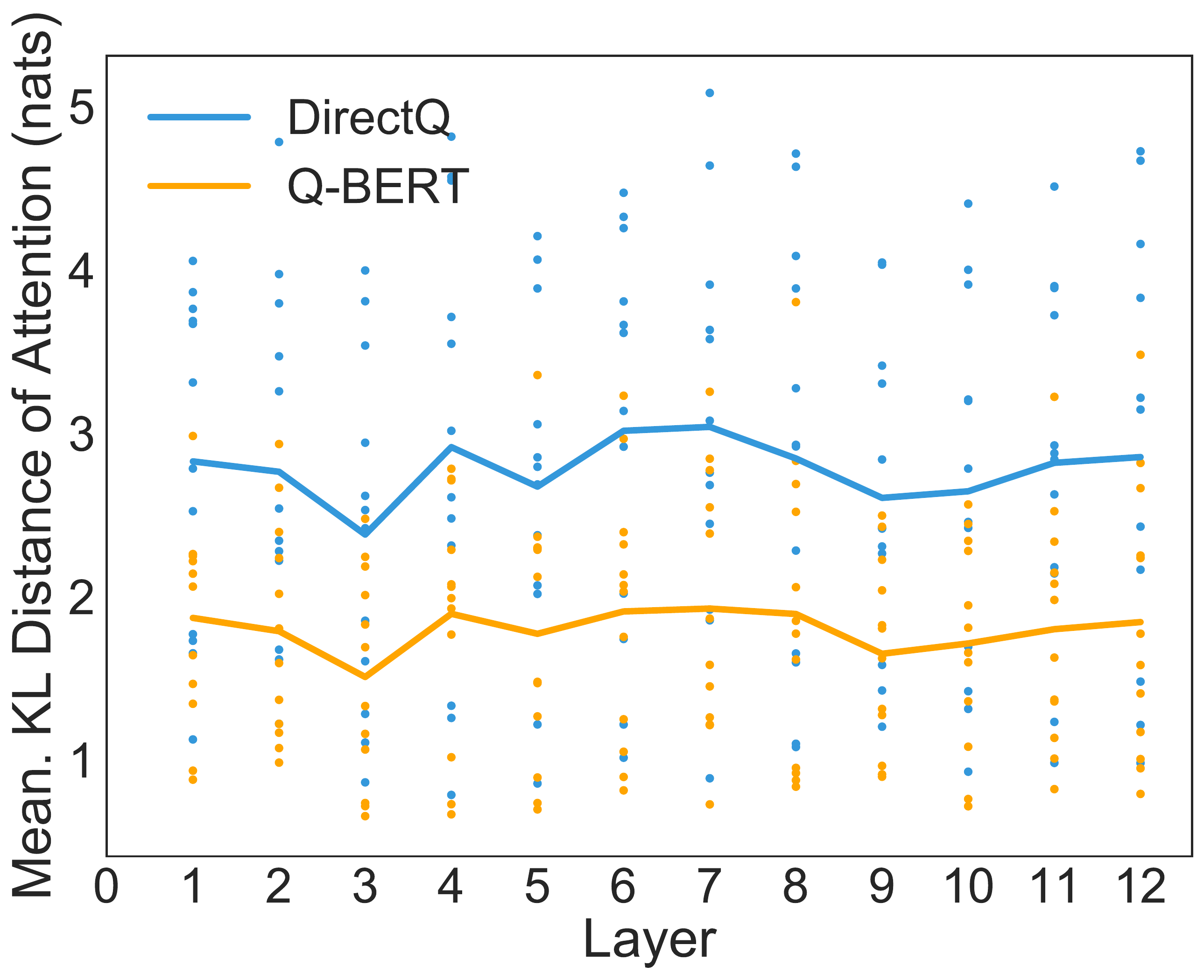}}
\end{center}
 \caption{\footnotesize
 KL divergence over attention distribution between \OURS/DirectQ and 
 Baseline. The distance between \OURS~and Baseline is much 
 smaller than that of DirectQ and Baseline.}
 \label{fig:attn_KL}
\end{figure}

To do so, we compute the Kullback–Leibler (KL) divergence
between the attention distribution for the same input from the coordinated
head of both quantized BERT and full-precision BERT.
It should be noted that we compute the average distance out of 10\% of the entire training dataset.
The smaller KL divergence here means that the output of the multi-head attention
of the two models is closer to each other.
We illustrate this distance score for each individual head in~\fref{fig:attn_KL} for SST-2, MNLI, CoNLL-03 and SQuAD.
We compared \OURS~and DirectQ with 4-bits weights, 8-bits embedding and 8-bits
activation. Each scatter point in~\fref{fig:attn_KL}
denotes the distance w.r.t. one head, and the line chart shows the average results over
the 12 heads in one layer. 
We can clearly see that \OURS~always incurs a smaller distance to the original baseline
model as compared to DirectQ model, for all different layers.

\section{Conclusion}

In this work, we perform an extensive analysis of fine-tuned BERT and propose \OURS, an effective scheme for quantizing BERT. 
In order to aggressively reduce the model size by mixed-precision quantization, we proposed a new layer-wise Hessian based
method which captures both the average and the variance of the eigenvalues.  
Moreover, a new group-wise quantization is proposed to perform fine-grained quantization inside each encoder layer.
In four downstream tasks, equipped with the aforementioned methods, \OURS~achieves $13\times$ compression ratio in weights, 
$4\times$ smaller activation size, and $4\times$ smaller embedding size, with at most 2.3\% accuracy loss.
To better understand how different factors will affect the trade-off between performance and the model compression ratio
in \OURS, we conduct controlled experiments to investigate the effect of different quantization schemes 
and quantizing different modules in BERT, respectively.

\section*{Acknowledgments}
We would like to thank Prof. Joseph Gonzalez, Prof. Dan Klein, and Prof. David Patterson for
their valuable feedback.
This work was supported by a gracious fund from Intel corporation,
Berkeley Deep Drive (BDD), and Berkeley AI Research (BAIR) sponsors.
We would like
to thank the Intel VLAB team for providing us with access to their computing cluster.
We also thank gracious support from Google for providing cloud compute.
MWM would also 
like to acknowledge ARO, DARPA, NSF, ONR, and Intel for providing partial support of this work.

\clearpage
\printbibliography

\clearpage
\onecolumn
\appendix

\begin{algorithm}[t]
\DontPrintSemicolon
\caption{Power Iteration for Eigenvalue Computation}
\label{alg:power_iteration}
    \SetAlgoLined
    \KwInput{Block Parameter: $W_i$.
    }
    
    Compute the gradient of $W_i$ by backpropagation, \emph{i.e.}, $g_i=\frac{d L}{d W_i}$.
    
    Draw a random vector $v$  (same dimension as $W_i$).
    
    Normalize $v$, $v=\frac{v}{\|v\|_2}$
    
    \For(\ \ \quad \quad\quad\quad\quad\tcp*[h]{Power Iteration}){i $=1,2,\ldots, n$}{
        Compute $gv = g_i^Tv$ \tcp*{Inner product}
        
        Compute $Hv$ by backpropagation, $Hv = \frac{d(gv)}{dW_i}$ \tcp*{Get Hessian vector product}
        
        Normalize and reset $v$, $v = \frac{Hv}{\|Hv\|_2}$
    }
\end{algorithm}
\section{Detailed quantization process}\label{sec:appendix_quantizaiton_process}
In the forward pass, each element in a weight or activation tensor $X$ will be quantized as~follows:
\begin{align*}
X^\prime = \text{Clamp}&(X, q_0, q_{2^{k}-1}), \\
X^I = \lfloor \frac{X^\prime - q_0}{\Delta} \rceil,& \text{ where } \Delta = \frac{q_{2^{k}-1} - q_0}{2^k - 1}, \\
Q(X) = \Delta &X^I + q_0 \\
\end{align*}
where $\lfloor \cdot \rceil$ is the round operator, $\Delta$ is the distance between adjacent quantized points, $X^I$ is a set of integer indices and so as $\text{bias}^I$, which is omitted for the clarity
in following equations. $[q_0, q_{2^{k}- 1}]$ stands for the quantization range of the floating point tensor, and the function Clamp sets all elements smaller than $q_0$ equal to $q_0$, and elements larger than $q_{2^{k}- 1}$ to $q_{2^{k}- 1}$. It should be noted that $[q_0, q_{2^{k}- 1}]$ can be a subinterval of $[min, max]$, in order to get rid of outliers and better represent the majority of the given tensor. 
During inference, the expensive floating point arithmetic can be replaced 
by efficient integer arithmetic for the matrix multiplication with $X^I$, 
and then followed by a gathered dequantization operation, 
which will significantly accelerate the computation process. Since we use the quantization-aware fine-tuning scheme, in the backward pass, the Straight-Through Estimator (STE)~\cite{bengio2013estimating} is used for computing the gradient for $X$.

\section{Dataset}\label{sec:appendix_dataset}
We apply \OURS~on Sentiment Classification, Natural Language Inference, 
Named Entity Recognition and Machine Reading Comprehension tasks. For 
Sentiment Classification, we evaluate on Stanford Sentiment Treebank (SST-2) \cite{socher2013recursive}. 
For Named Entity Recognition, we use CoNLL-2003 English benchmark dataset for NER (CoNLL-03) \cite{sang2003introduction}. 
For Natural Language Inference, we test on Multi-Genre Natural Language Inference (MNLI) \cite{williams2017broad}. 
For Machine Reading Comprehension, we evaluate on the Stanford Question Answering Dataset (SQuAD) \cite{rajpurkar2016SQuAD}. 

More specifically, SST-2 is a movie review dataset with binary annotations, 
where the binary label indicates positive and negative reviews. 
MNLI is a multi-genre NLI task for predicting whether a given premise-hypothesis 
pair is entailment, contradiction or neural. Its test and development
datasets are further divided into in-domain (MNLI-m) and cross-domain 
(MNLI-mm) splits to evaluate the generality of tested models. CoNLL-03 
is a newswire article dataset for predicting the exact span of the annotated 
four entity types: person, location, organization, and miscellaneous. 
SQuAD is a task to answer the question by extracting the relevant span 
from the context, where a paragraph of context and a question is provided for each sample. 

\section{Extra results}
Here we describe several additional results.

\subsection{Ablation Study of Hessian based Mixed Precision Assignment}\label{sec:ab-hessian-bits}
To demonstrate the effectiveness of our Hessian based Mixed Precision method, 
we conduct the ablation study here to use the reversed version of 2/3-bit \OURS\MP ~(\OURS\MP-rev).
Specifically, we will assign lower bits to relatively sensitive layers 
and higher bits vice versa while keeping model size the same. This means the previous layer in 2/3-bit \OURS\MP~with 2-bit will be assigned 3-bit in \OURS\MP-rev. \footnote{The bits setting of 2/3-bit \OURS\MP~and 2/4-bit \OURS\MP~are included in~\tref{tab:bits_st} and ~\tref{tab:bits-setting-24}, respectively.}

We can obverse that with the same model size, the performance difference 
between \OURS\MP-rev and 2-bit \OURS~is within 2\% for MNLI, CoNLL-03, SQuAD 
and 4\%~for SST-2, while that of \OURS\MP~is beyond 5\% for MNLI, CoNLL-03, SQuAD and 8\% for SST-2. 
This large discrepancy in the performance illustrates the superiority of leveraging second order Hessian
information in mix precision bits assignment.

\begin{table}[!htbp]
    \caption{
    \footnotesize
    Quantization results for reversed Hessian based Mixed Precision setting. We denote the model using reversed bit assignment against \OURS\MP~as \OURS\MP-rev.
    }
\centering
\subfloat[SST-2]{
\centering
\begin{tabular}{lcccccccccccccc}
\toprule
Method & w-bits & e-bits & Acc\\
\midrule
Baseline & 32 & 32 & 93.00  \\
\midrule
\OURS & 2 & 8 & {84.63} \\
\midrule
\OURS\MP-rev & 2/3~\MP & 8 & {88.42} \\
\midrule
\hc \OURS\MP & 2/3~\MP & 8 & \textbf{92.08} \\
\bottomrule 
\end{tabular}
\label{tab:sst-rv}
}
\subfloat[MNLI]{
\centering
\begin{tabular}{lcccccccccccccc}
\toprule
Method & w-bits & e-bits & Acc-m & Acc-mm \\
\midrule
Baseline & 32 & 32 & 84.00 & 84.40 \\
\midrule
\OURS & 2 & 8 & {76.56} & {77.02}\\
\midrule
\OURS\MP-rev & 2/3~\MP & 8 & {78.91} & {79.30}\\
\midrule
\hc \OURS\MP & 2/3~\MP & 8 & \textbf{81.75} & \textbf{82.29}\\
\bottomrule
\end{tabular}
\label{tab:mnli-rv}
}

\subfloat[CoNLL-03]{
\centering
\begin{tabular}{lccccccccccc}
\toprule
Method & w-bits & e-bits & F$_1$\\
\midrule
Baseline & 32 & 32 & 95.00 \\
\midrule
\OURS & 2 & 8 & {91.06} \\
\midrule
\OURS\MP-rev & 2/3~\MP & 8 & {92.66} \\
\midrule
\hc \OURS\MP & 2/3~\MP & 8 & \textbf{94.37} \\
\bottomrule 
\end{tabular}
\label{tab:ner-rv}
}
\subfloat[SQuAD]{
\begin{tabular}{lccccccccccc}
\toprule
Method & w-bits & e-bits & EM & F$_1$ \\
\midrule
Baseline & 32 & 32 & 81.54 & 88.69 \\
\midrule
\OURS & 2 & 8 & {69.68} & {79.60}\\
\midrule
\OURS\MP-rev & 2/3~\MP & 8 & {69.71} & {79.39}\\
\midrule
\hc \OURS\MP & 2/3~\MP & 8 & \textbf{79.29} & \textbf{86.95}\\
\bottomrule 
\end{tabular}
\label{tab:SQuAD-rv}
}
\label{tab:main-rv}
\end{table}

\vspace{-3mm}
\subsection{Mixed Precision Quantization for Embedding}\label{sec:emb-mix-quantize}
As can be seen from~\tref{tab:main}, when 2/3~\MP ~is used for quantizing the weight parameters, 
the bottleneck of the model size is bounded by the 
embedding table size. 
Also, observed in~\tref{tab:emb}, we noticed that word embedding is less sensitive.
Therefore, in this section, we further push the embedding table to be 4-bit (word embedding) 
and 8-bit (position embedding) mixed-precision to reduce the entire model size. 
Similar to group-wise quantization for weights, in this ultra-low embedding bits setting, 
we bucket the 768 output neurons in \bertbase~word and position embedding layer into 128 groups in~\tref{tab:main-mix-emb}.
We adopt the same setting for weights and activations in ~\tref{tab:main}, 
where we employ 128 groups for weights and set 8 bits for activation.
Note that with around 0.5\% performance drop, the embedding table size can be reduced to 11.6MB, which corresponds to around $8\times$ compression ratio in embedding table and $12\times$ compression ratio in total model size.

\vspace{-8mm}
\begin{table}[!htbp]
    \caption{
    \footnotesize
    Embedding mixed-precision quantization results for \OURS~on four tasks. 
    Results are obtained with 128 groups in each encoder layer and \textbf{embedding layer}.
    We abbreviate quantization bits used for weights as ``w-bits'', embedding as ``e-bits'', model size in MB as ``Size'', and model size without embedding layer in MB as ``Size-w/o-e''. For simplicity and efficacy, all the models except for Baseline are using 8\text{-bits} activation. Here ``MP'' refers to mixed-precision quantization. The mixed-precision embedding here uses 4-bit word embedding and 8-bit position embedding.
    }
    \label{tab:main-mix-emb}
\centering
\setlength\tabcolsep{2.35pt}
\subfloat[\footnotesize SST-2]{
\centering
\begin{tabular}{lcccccccccccccc}
\toprule
Method      & w-bits    & e-bits    & Acc           & Size      & Size-w/o-e\\
\midrule        
Baseline    & 32        & 32        & 93.00         & 415.4     & 324.5 \\
\midrule 
\hc\OURS\MP & 2/4~\MP   & 8         & \bf{92.55}    & 53.2      & 30.5      \\
\hb\OURS\MP & 2/4~\MP   & 4/8~\MP   & \bf{92.32}      & 42.0      & 30.5      \\
\midrule                        
\hc\OURS\MP & 2/3~\MP   & 8         & \bf{92.08}    & \bf{48.1} & \bf{25.4} \\
\hb\OURS\MP & 2/3~\MP   & 4/8~\MP   & \bf{91.51}    & \bf{36.9} & \bf{25.4} \\
\bottomrule 
\end{tabular}
\label{tab:sst-mix-emb}
}
\subfloat[\footnotesize MNLI]{
\centering
\begin{tabular}{lcccccccccccccc}
\toprule
Method          & w-bits    & e-bits    & Acc-m          & Acc-mm      &   Size    &   Size w/o-e\\
\midrule
Baseline        & 32        & 32        & 84.00         & 84.40     & 415.4   &   324.5 \\
\midrule 
\hc \OURS\MP    & 2/4~\MP   & 8         & \bf{83.51}    & \bf{83.55}    & 53.2 & 30.5\\
\hb \OURS\MP    & 2/4~\MP   & 4/8~\MP & \bf{82.82}    & \bf{83.36}    & 42.0 & 30.5\\
\midrule 
\hc \OURS\MP    & 2/3~\MP   & 8         & \bf{81.75}    & \bf{82.29}    & \bf{46.1} & \bf{23.4}\\
\hb \OURS\MP    & 2/3~\MP   & 4/8~\MP & \bf{81.00}    & \bf{81.65}    & \bf{34.9} & \bf{23.4}\\
\bottomrule
\end{tabular}
\label{tab:mnli-mix-emb}
}

\vspace{-1mm}
\subfloat[\footnotesize CoNLL-03]{
\centering
\begin{tabular}{lcccccccccccccc}
\toprule
Method          & w-bits    & e-bits    & F$_1$         & Size      & Size-w/o-e    \\
\midrule                    
Baseline        & 32        & 32        & 95.00         & 410.9     & 324.5         \\
\midrule                                
\hc \OURS\MP    & 2/4~\MP   & 8         & \bf{94.55}    & 52.1      & 30.5          \\ 
\hb \OURS\MP    & 2/4~\MP   & 4/8~\MP   & \bf{94.55}    & 41.5      & 30.5          \\ 
\midrule                                    
\hc \OURS\MP    & 2/3~\MP   & 8         & \bf{94.37}    & \bf{45.0} & \bf{23.4}     \\ 
\hb \OURS\MP    & 2/3~\MP   & 4/8~\MP   & \bf{94.45}    & \bf{34.4} & \bf{23.4}     \\ 
\bottomrule 
\end{tabular}
\label{tab:ner-mix-emb}
}
\vspace{-1mm}
\subfloat[\footnotesize SQuAD]{
\begin{tabular}{lccccccccccccccc}
\toprule
Method          & w-bits    & e-bits    & EM            & F$_1$         & Size      & Size-w/o-e\\
\midrule                    
Baseline            & 32            & 32        & 81.54         & 88.69         & 415.4     & 324.5     \\
\midrule                                
\hc \OURS\MP        & 2/4~\MP     & 8   & \bf{79.85}    & \bf{87.49}    & 53.2      & 30.5      \\
\hb \OURS\MP        & 2/4~\MP       & 4/8~\MP   & \bf{79.53}    & \bf{87.14}    & 42.0      & 30.5      \\
\midrule                                    
\hc \OURS\MP        & 2/3~\MP       & 8         & \bf{79.29}    & \bf{86.95}    & \bf{48.1} & \bf{25.4} \\
\hb \OURS\MP        & 2/3~\MP       & 4/8~\MP   & \bf{78.68}    & \bf{86.49}        & \bf{36.9} & \bf{25.4} \\
\bottomrule 
\end{tabular}
\label{tab:squad-mix-emb}
}
\end{table}

\subsection{Detailed loss landscape for SST-2}
\begin{figure}[!htp]
\begin{center}
 \subfloat[SST-2 $3^\text{th}$ layer ]{\label{subfig:sst-loss-1}\includegraphics[width=.4\linewidth]{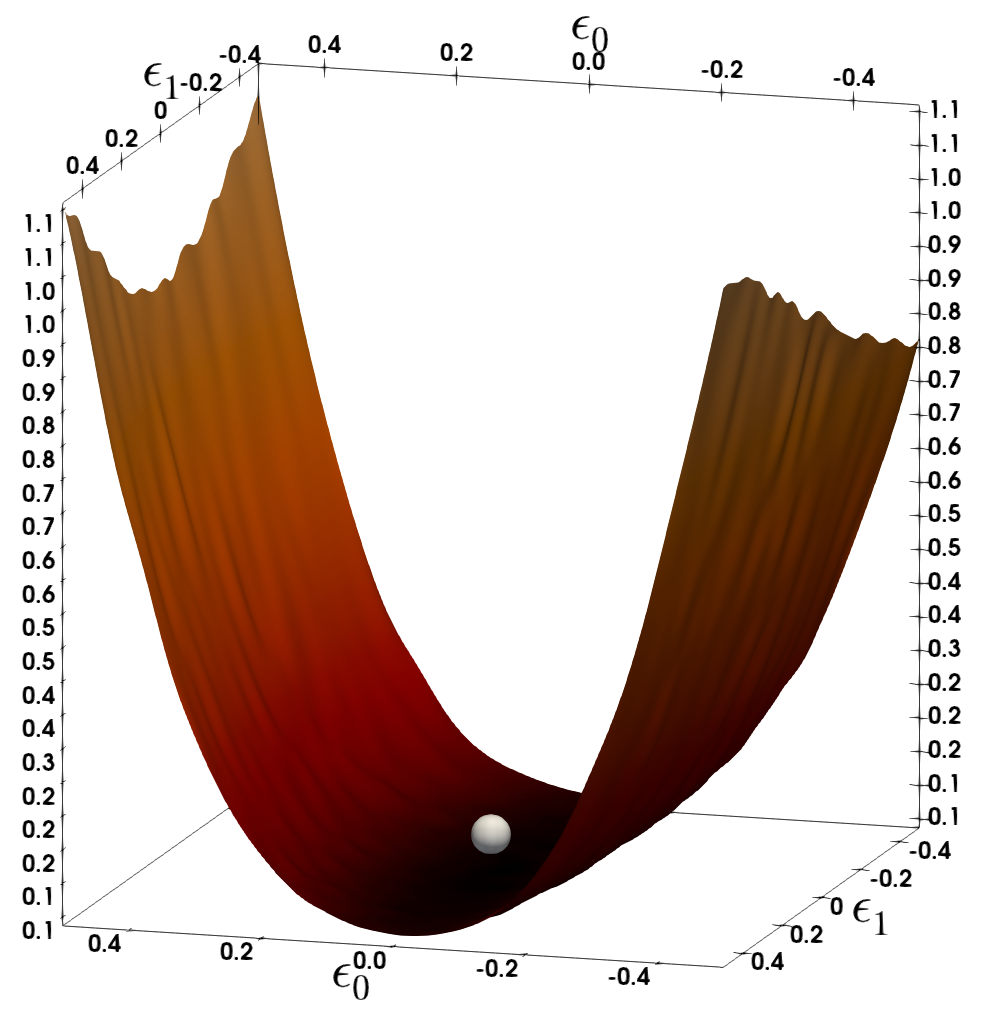}}
  \subfloat[SST-2 $10^\text{th}$ layer ]{\label{subfig:sst-loss-2}\includegraphics[width=.4\linewidth]{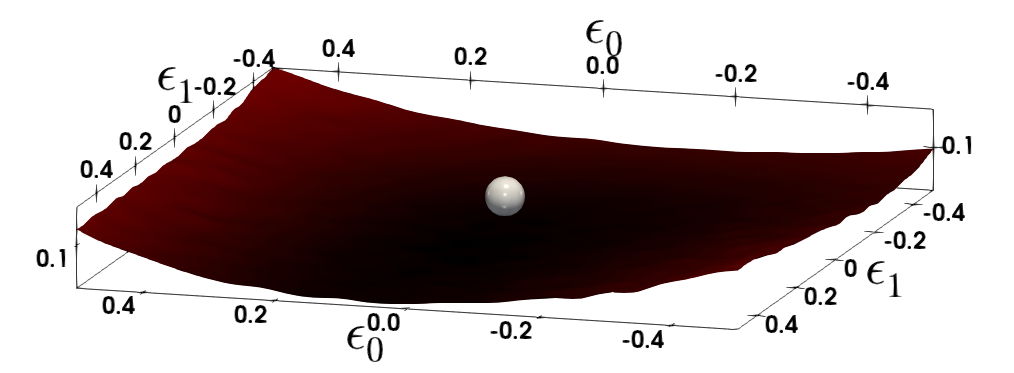}}
\end{center}
 \caption{\footnotesize
The loss landscape for different layers in SST-2 is
illustrated by perturbing the parameters
along the first two dominant eigenvectors of the Hessian. 
The silver sphere shows the point in the parameter space that the BERT model
has converged to.
 }
\label{fig:Hessian-loss-landscape-sst2}
\end{figure}

We include the detailed loss landscape analysis for the remaining task SST-2 as shown in~\fref{fig:Hessian-loss-landscape-sst2}.

\centering

\begin{table}[!htbp]
\caption{Bits setting for 2/3-bit \OURS\MP~in all four tasks.}
\small
\setlength\tabcolsep{2.pt}
\label{tab:bits_st}
\centering
\begin{tabular}{lcccccccccccccc} \toprule
\ha &Layer(s)      &Layer Type    &Parameter Size(M)    & Weight bit (SST-2)    & Weight bit (MNLI)  & Weight bit (CoNLL-03) &   Weight bit (SQuAD) \\
\midrule
\hc & Layer 0   & Embedding     &   23.8         &     8     &     8   &   8   &   8   \\
\midrule
\ha & Layer 1   & Transformer   &   7.1        &    2               &     2          &     2         &   2 \\
\midrule
\hc & Layer 2   & Transformer   &   7.1         &    2               &     2          &     3         &   2 \\
\midrule
\ha & Layer 3   & Transformer   &   7.1         &    2               &     2          &     3         &   2 \\
\midrule
\hc & Layer 4   & Transformer   &   7.1         &    3               &     2          &     3         &   3 \\
\midrule
\ha & Layer 5   & Transformer   &   7.1         &    3               &     3          &     3         &   3 \\
\midrule
\hc & Layer 6  & Transformer    &   7.1        &    3               &     3          &     2         &   3 \\
\midrule
\ha & Layer 7 & Transformer     &   7.1         &    3               &     3          &     2         &   3 \\
\midrule
\hc & Layer 8  & Transformer    &   7.1         &    3               &     3          &     2         &   3 \\
\midrule
\ha & Layer 9  & Transformer    &   7.1         &    3               &     2          &     2         &   3 \\
\midrule
\hc & Layer 10  & Transformer   &   7.1         &    2               &     2          &     2         &   2 \\
\midrule
\ha & Layer 11  & Transformer   &   7.1         &    2               &     2          &     2         &   2 \\
\midrule
\hc & Layer 12 & Transformer    &   7.1         &    2               &     2          &     2         &   2  \\
\midrule
\ha & Layer 13 & FC             &   0.01        &    32              &     32         &     32        &   32   \\
\bottomrule
\end{tabular}
\end{table}

\begin{table}[!htbp]
\caption{Bits setting for 2/4-bit \OURS\MP~in all four tasks.}
\small
\setlength\tabcolsep{2.pt}
\label{tab:bits-setting-24}
\centering
\begin{tabular}{lcccccccccccccc} \toprule
\ha &Layer(s)      &Layer Type    &Parameter Size(M)    & Weight bit (SST-2)    & Weight bit (MNLI)  & Weight bit (CoNLL-03) &   Weight bit (SQuAD) \\
\midrule
\hc & Layer 0   & Embedding     &   23.8         &     8     &     8   &   8   &   8   \\
\midrule
\ha & Layer 1   & Transformer   &   7.1        &    2               &     2          &     2         &   2 \\
\midrule
\hc & Layer 2   & Transformer   &   7.1         &   2               &     2          &     2         &   2 \\
\midrule
\ha & Layer 3   & Transformer   &   7.1         &    4               &     2          &     2        &   2 \\
\midrule
\hc & Layer 4   & Transformer   &   7.1         &    4               &     4          &     4         &   4 \\
\midrule
\ha & Layer 5   & Transformer   &   7.1         &    4               &     4          &     4         &   4 \\
\midrule
\hc & Layer 6  & Transformer    &   7.1        &    2               &     4          &     4         &   4 \\
\midrule
\ha & Layer 7 & Transformer     &   7.1         &    4               &     4          &     4         &   4 \\
\midrule
\hc & Layer 8  & Transformer    &   7.1         &    4               &     4          &     4         &   4\\
\midrule
\ha & Layer 9  & Transformer    &   7.1         &    4               &     2          &     2         &   2 \\
\midrule
\hc & Layer 10  & Transformer   &   7.1         &    2               &     2          &     2         &   2 \\
\midrule
\ha & Layer 11  & Transformer   &   7.1         &    2               &     2          &     2         &   2 \\
\midrule
\hc & Layer 12 & Transformer    &   7.1         &    2               &     2          &     2         &   2  \\
\midrule
\ha & Layer 13 & FC             &   0.01        &    32              &     32         &     32        &   32   \\
\bottomrule
\end{tabular}
\end{table}




\end{document}